\documentclass[journal]{IEEEtran}
\usepackage{cite}

\ifCLASSINFOpdf
  \usepackage[pdftex]{graphicx}
\else
  \usepackage[dvips]{graphicx}
\fi
\usepackage{amsmath}
\usepackage{algorithm}
\usepackage{algorithmic}

\usepackage{flushend}
\usepackage{xcolor}
\usepackage{subcaption}
\usepackage{bbold}

\begin{document}
%
\title{Small-Footprint Open-Vocabulary Keyword Spotting with Quantized LSTM Networks}
%
%
%

\author{Th{\'e}odore Bluche, Ma{\"e}l Primet, Thibault Gisselbrecht\\
Sonos Inc., Paris, France
}
\newcommand{\todo}[1]{\colorbox{orange!30}{\color{red}\textbf{TODO:} #1}}
\newcommand{\x}{\mathbf{x}}
\newcommand{\y}{\mathbf{y}}
\newcommand{\interm}{\mathbf{z}}
\newcommand{\pseq}{\mathbf{\pi}}
\newcommand{\labseq}{\mathbf{l}}
\maketitle

\begin{abstract}
We explore a keyword-based spoken language understanding system, in which the intent of the user can directly be derived from the detection of a sequence of keywords in the query. In this paper, we focus on an open-vocabulary keyword spotting method, allowing the user to define their own keywords without having to retrain the whole model. We describe the different design choices leading to a fast and small-footprint system, able to run on tiny devices, for any arbitrary set of user-defined keywords, without training data specific to those keywords. The model, based on a quantized long short-term memory~(LSTM) neural network, trained with connectionist temporal classification~(CTC), weighs less than 500KB. Our approach takes advantage of some properties of the predictions of CTC-trained networks to calibrate the confidence scores and implement a fast detection algorithm. The proposed system outperforms a standard keyword-filler model approach.
\end{abstract}

%
\IEEEpeerreviewmaketitle

\section{Introduction}
\label{sec:intro}
%
%
%
%

\IEEEPARstart{A}{utomatic} speech recognition~(ASR) systems have recently reached close to human recognition performance~\cite{xiong2016achieving}, allowing voice assistants~(Alexa, Google Assistant, Siri), vocal interfaces and other spoken language understanding~(SLU) systems to flourish. However, to achieve such performance, most \textit{``ask-me-anything''} voice assistants run large-vocabulary continuous speech recognition~(LVCSR) models, demanding a lot of resource and computing power. Therefore, most of the processing is performed in the cloud, inducing privacy concerns and latency issues. When SLU is limited to a specific number of tasks, in a closed-ontology setting (e.g. with a task-specific language model~\cite{saade2018spoken}), the inference can be performed on device. Recently, generic ASR models running on mobile devices have also been proposed~\cite{he2019streaming}. In both cases, the full systems weigh more than 100MB, which remains too large for small devices typical of IoT applications where the memory and computing power is scarce. 

\begin{figure}[!t]
    \centering
    \includegraphics[width=\linewidth]{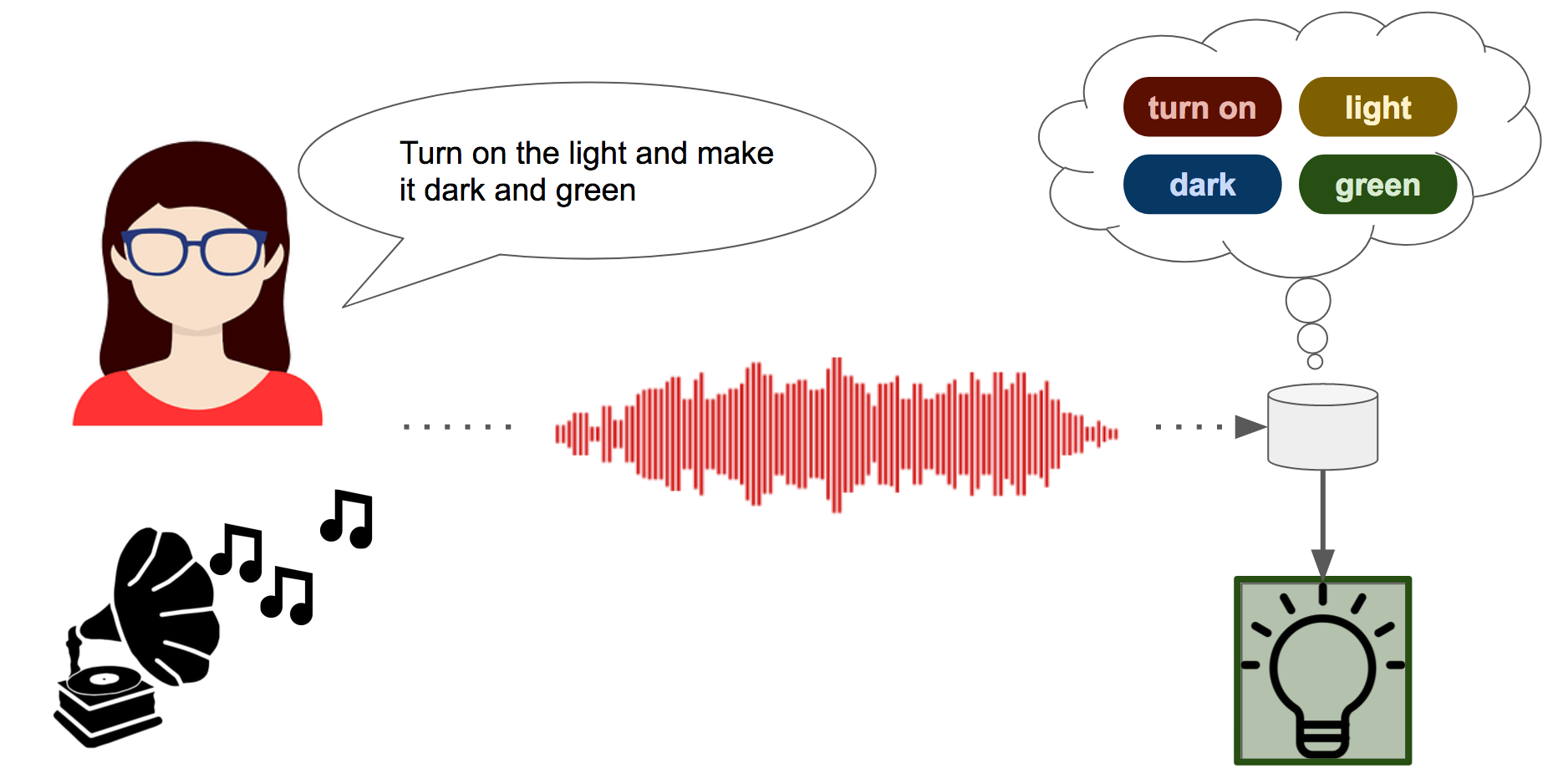}
    \caption{Mini-SLU system based on keyword spotting. The user says a query in natural language. The system performs an action based on the detection of keywords in the query.}
    \label{fig:mini_asr}
\end{figure}

We target ``mini-SLU'' scenarios, in which the detection of simple keywords in the query is sufficient to convey its meaning. In such a system, the user should be able to speak in natural language to trigger an action based on the keywords, as illustrated on Fig.~\ref{fig:mini_asr}. For this system to be practical and easy to adapt to any use-case, we assume that it should adapt to situations where the set of keywords is not known in advance, allowing the user to define their own interactions based on custom keywords. This also implies that no specific training data is available.

We present in this paper a keyword spotting~(KWS) system, designed to be small enough to fit on micro-controllers, i.e. weigh less than 500KB. The development of tiny KWS models is an active area of research, mainly focusing on the detection of wake words or a pre-defined set of commands allowing single-word interactions. For these applications, it is reasonably feasible to collect a training dataset labeled at the keyword level (e.g. the Google Speech Commands~\cite{warden2018speech} or ``Hey Snips''~\cite{coucke2019efficient} datasets). These works focus mainly on the neural network architecture (feed-forward~\cite{chen2014small}; convolutional~\cite{sainath2015convolutional}; residual~\cite{tang2018deep}; recurrent~\cite{fernandez2007application,sun2016max} neural networks; WaveNet~\cite{coucke2019efficient}), or on the compression methods~\cite{tucker2016model,zhang2017hello,amoh2019optimized}. The networks are usually trained at the frame level using the cross-entropy loss. Other choices of losses, such as the connectionist temporal classification (CTC)~\cite{graves2006connectionist,fernandez2007application} or a max-pooling loss~\cite{sun2016max} have also been proposed. Although they have an attractive formulation, since the neural network directly predicts the confidence at the keyword level, these methods are not suited to the scenario we explore, because they require to know the set of keywords in advance, and a specific training set made of these keywords.

Historically, the approach consisting in modeling the keyword directly to score segments of audio~\cite{bridle1973efficient,wilpon1990automatic} evolved into acoustic KWS, mainly based on hidden Markov models~(HMMs). These methods take advantage of the modeling of sub-word units (e.g. phone) by the HMMs to enable building acoustic models of any arbitrary keyword, only requiring generic ASR training data. To cope with the issue of scoring and comparing acoustic segments of different lengths, these approaches generally involve a \textit{``filler model''} of speech segments outside the keyword~\cite{rohlicek1989continuous,rose1990hidden,weintraub1993keyword,szoke2005comparison} (e.g. an ergodic phone HMM). A background model may be applied to compute the likelihood ratio between keyword and generic speech~\cite{weintraub1995lvcsr}. A survey on acoustic KWS can be found in~\cite{mandal2014recent}.

Similarly to what has been done in ASR, HMMs have been replaced
with neural networks that predict the phone or grapheme posteriors directly, for example with CTC training~\cite{hwang2015online,lengerich2016end,chen2018sequence}. Since the network predicts phone posteriors, the filler model may either be ignored, because it will always give a probability of one, replaced by the greedy prediction of the network~\cite{hwang2015online}, or augmented with a phone language model~\cite{he2017streaming}. When the neural network is very small, it tends to make phoneme or grapheme prediction errors. The systems can take advantage of the network predicting phonemes or graphemes by augmenting the keyword set with alternative pronunciations, either estimated from the training set~\cite{yang2018automatic} or from examples spoken by the user~\cite{lugosch2018donut}. Using the knowledge about the confusions of the network along with the peaky behaviour of CTC-trained networks, an efficient detection can be implemented based on a minimum edit distance search of the keywords in a compact phone lattice~\cite{zhuang2016unrestricted}. 

Recently, a few end-to-end neural networks for open-vocabulary KWS have been proposed, that rely on the embedding in a vector space of the audio on the one hand, and of the keyword phone sequence on the other hand, followed by a detection decided based on the distance between the two~\cite{sacchi2019open} or by a neural network~\cite{audhkhasi2017end}. However, these last two methods seem to be only applicable to single-keyword queries, which do not fit our mini-SLU scenario.

In this paper, we explore a method similar to~\cite{hwang2015online}, based on a CTC-trained neural network made of long short-term memory~(LSTM) layers. We particularly focus on making the system as small and fast as possible, to detect sequences of keywords in natural language. In particular, we compare different model sizes, choices of confidence scores, and optimizations of the decoding procedure. We evaluate the models on two crowd-sourced datasets for the task of mini-SLU. 
We compare the proposed approach to several baselines: LVCSR approaches based on Viterbi decoding and lattices, keyword-filler models, and other methods proposed for CTC-trained neural networks.
We show that we can have good results with models smaller than 500KB, outperforming the keyword-filler method. The final model can run in real time on micro-controllers.

The main contributions of this paper are the following:
\begin{itemize}
    \item We propose a quantization strategy for LSTM networks
    \item We devise a confidence score adapted to the particularities of the outputs of CTC-trained networks
    \item We propose a fast decoding strategy, which besides pruning, is made faster by skipping frames, using ideas similar to~\cite{chen2016phone,zhuang2016unrestricted}
    \item We carry out a comprehensive comparison of different design choices.
\end{itemize}

The remaining of the paper is as follows. In Section~\ref{sec:am} we present the acoustic model neural network: its architecture and how it is quantized and trained. We explain in Section~\ref{sec:asr-lite} our keyword spotting mechanism, including the confidence score settings and the various optimizations we explored. In Section~\ref{sec:setup} we describe the experimental setup, including the datasets and metrics we used. The results of our exploration are reported in Section~\ref{sec:results}.


\section{Acoustic Model}
\label{sec:am}

The method we present in this paper relies on a trained acoustic model. The main requirements for this model are: 
\textit{(i)} it has to be small, to fit on tiny devices such as micro-controllers, 
\textit{(ii)} it has to be compatible with streaming recognition, to enable real-time KWS, and
\textit{(iii)} it should be accurate enough for keywords to be detectable from its output. 

In this section, we present the chosen architecture. We build a multi-layer LSTM neural network~\cite{sak2015fast}, described in Section~\ref{sec:am-archi}. In order to keep the whole model under 500KB, we quantize the parameters and intermediate activations of the neural network. The chosen quantization scheme is presented in Section~\ref{sec:am-quantization}. The model is directly trained on a generic ASR dataset $\mathcal{D}$ with connectionist temporal classification~(CTC)~\cite{graves2006connectionist}, as presented in Section~\ref{sec:am-training}.

\subsection{Stack\&Skip LSTM architecture}
\label{sec:am-archi}

The inputs of the networks are sequences of stacks of $5$ consecutive MFCC frames, computed every $3$ frames~\cite{sak2015fast}.
The networks consist of a first affine layer with a $tanh$ activation, followed by a stack of LSTM layers.
The output $h_t$ of an LSTM layer at time $t$, for the input sequence $\mathbf{x} = x_1 x_2 \ldots x_T$ is computed as follows:
\begin{eqnarray}
    i_t & = & \sigma ( W_{ix} x_t + W_{ih} h_{t-1} + b_i ) \\
    j_t & = & tanh( W_{jx} x_t + W_{jh} h_{t-1} + b_j ) \\
    f_t & = & \sigma ( W_{fx} x_t + W_{fh} h_{t-1} + b_f ) \\
    c_t & = & f_t \odot c_{t-1} + i_t \odot j_t \\
    o_t & = & \sigma ( W_{ox} x_t + W_{oh} h_{t-1} + b_o ) \\
    h_t & = & o_t \odot tanh(c_t),
\end{eqnarray}
where $\{W_*, b_*\}$ are the free parameters of the LSTM, $i_t, f_t, o_t, j_t$ are respectively the input, forget, output gate and cell inputs, and $c_t$ is the internal state.
An affine output layer is added on top of the last LSTM to compute the class logits: one for each phone, plus one representing a \textit{``blank''} (or \textit{null}) class.

\subsection{Quantization Scheme}
\label{sec:am-quantization}

There has been some work on LSTM quantization, but either not explicit regarding the quantization scheme~\cite{mcgraw2016personalized,wang2018c} or only partially quantizing the LSTM for inference, with the internal state~\cite{he2016effective} or the whole LSTM~\cite{zen2016fast} kept in floating-point. In this section, we explain the quantization scheme we propose for LSTM layers.

All weights and activations are quantized to 8 bits, following a scheme similar to the one proposed by Jacod et al~\cite{jacob2018quantization}.
We use a special case of that scheme, with symmetric quantization ranges with power-of-two bounds. This choice simplifies the computation, thanks to the absence of offset and the changes of scale being implemented as bit shifts.

The weights are quantized post-training. The range is set to the next power of two to the maximum (or negative minimum) value of each weight matrix. To avoid the side effects of large outlier weights, the weights are first clipped to $[-8;+8]$, ensuring at least a precision of $\pm 0.0625$ once quantized.

The activations are quantized during training. Instead of computing the quantization range from min/max statistics, we use fixed ranges. This choice is motivated by several reasons. First, the LSTM contains saturating activation functions. Thus we know \textit{a priori} that their outputs will lie in $(-1;+1)$. Moreover, we may set a fixed range for their inputs, since large values will be in the saturating part anyway. We set that range to $(-4;+4)$. Using the same fixed range for all activation function inputs and outputs allows to have a single lookup table at inference for the sigmoid and the tanh functions, making the model faster to execute. The second motivation comes from the fact that the LSTM contains many additions, which are easier to implement if the operands have the same quantization parameters. Finally, the inner state $c_t$ is not bounded, since it can increase by one at each time step. If it were quantized using min/max statistics, we might loose at lot of precision.

The equations of the computation of the LSTM inner state and output are modified as follows:
\begin{eqnarray}
    i_t & = & Q_1[ \sigma ( Q_4[ W_{ix} x_t + W_{ih} h_{t-1} + b_i ] ) ] \\
    j_t & = & Q_1[ tanh( Q_4[ W_{jx} x_t + W_{jh} h_{t-1} + b_j ] ) ] \\
    f_t & = & Q_1[ \sigma ( Q_4[ W_{fx} x_t + W_{fh} h_{t-1} + b_f ] ) ] \\
    c_t & = & Q_4[ f_t \odot c_{t-1} + i_t \odot j_t ] \\
    o_t & = & Q_1[ \sigma ( Q_4[ W_{ox} x_t + W_{oh} h_{t-1} + b_o ] ) ] \\
    h_t & = & Q_1[ o_t \odot Q_1[ tanh(c_t) ] ],
\end{eqnarray}
where $Q_r$ represents the quantization of the values in the range $[-r;+r)$. During training, we use floating-point quantized values with the following \textit{fake quantization} operator:
\begin{eqnarray}
    \tilde{Q_r}(v) & = & \lceil  v \times 128 / r \rfloor \\
    Q_r(v) & = & clamp(\tilde{Q_r}(v), -128, 127) \times r / 128,
\end{eqnarray}
where $\lceil \cdot \rfloor$ is the rounding operation.

\subsection{Training}
\label{sec:am-training}

The acoustic model is trained with the CTC loss: an end-to-end training method that does not require to align the data prior to training. The goal is to minimize the following loss:
\begin{equation}
    \mathcal{L}_{CTC} = - \sum_{(\x^{(i)}, \pseq^{(i)}) \in \mathcal{D}} \log p(\pseq^{(i)} | \x^{(i)}),
\end{equation}
where, $\mathcal{D} = \{(\x^{(i)}, \pseq^{(i)})\}$ is a dataset of audio feature vector sequences with the corresponding phone target sequences.

To deal with the fact that the phone sequence $\pseq$ and inputs $\x$ have different lengths, the CTC extend the phone alphabet $\mathcal{P}$ with a so-called ``\textit{blank}'' class $\oslash$: $\mathcal{P}' = \mathcal{P} \cup \{\oslash\}$ and defines a simple mapping $\mathcal{B}: \mathcal{P}^{'*} \mapsto \mathcal{P}^*$ that removes symbol repetitions and blanks. For example:
\begin{equation*}
    \mathcal{B}(a~a~\oslash~b~b~b~\oslash~\oslash~c) = a~b~c.
\end{equation*}
Then we can build the set of all label sequences of a given length $T$ that yield a given phone sequence through $\mathcal{B}$:
\begin{equation*}
    \mathcal{B}^{-1}_T(\pseq) = \{\mathbf{l} = l_1 l_2 \ldots l_T : \mathcal{B}(\mathbf{l}) = \pseq \}.
\end{equation*}
With a conditional independence assumption on the labels, the posterior phone sequence probability can be rewritten as 
\begin{equation}
    p(\pseq | \x) = \sum_{\l \in \mathcal{B}^{-1}_T(\pseq)} \prod_{t=1}^{T} p(l_t | \x),
\end{equation}
where $p(l_t | \x)$ corresponds to the $l_t$-th output of the acoustic model at time $t$, allowing to compute and minimize the CTC loss with gradient descent.

\begin{table}[!t]
    \begin{center}
    \begin{tabular}{rrccc}
         \textbf{Layers} & \textbf{Units} & \textbf{Num. Params} & \textbf{Model size} & \textbf{Quantized size}  \\\hline
         3 &  64 & 115k & 460kB & 115kB \\
         5 &  64 & 181k & 724kB & 181kB \\
         3 &  96 & 246k & 984kB & 246kB \\
         5 &  96 & 395k & 1.5MB & 395kB  \\
         3 & 128 & 427k & 1.7MB & 427kB \\\hline
         \multicolumn{2}{r}{TDNN-LSTM} & 2.6M & 10.5MB & - \\\hline
    \end{tabular}
    \end{center}
    \caption{Size of the base acoustic models evaluated in this paper, with different number of layers and of units on each layer. The quantized networks use one byte per parameter.}
    \label{tab:nnet-sizes}
\end{table}

The weights are quantized after training.
To make sure that the quantization range is not too big, and to keep enough precision of the weights, we add an $L_2$-regularization loss, with a weight decay parameter of $0.0005$. The quantized LSTM implementation is quite slow compared to the cuDNN LSTM implementation. Therefore we start by training a model without quantization for 40 epochs using the cuDNN implementation. We then activate the fake quantization of the activations and train for an additional four epochs. 

To measure the impact of the size of the acoustic model, we train several neural networks of different sizes (Table~\ref{tab:nnet-sizes}).
All tested networks have less than 500k parameters, and weigh less than 500kB in their quantized form. Very small networks with 115 to 250k parameters were also evaluated, as they can be compared in size to modern KWS neural networks trained with keyword-specific datasets~\cite{chen2014small,sainath2015convolutional}. Finally, we also compare these models to a time-delay and LSTM neural network (TDNN-LSTM) hybrid NN/HMM model with $ 1,600$ tied biphone states trained with Kaldi with the lattice-free MMI objective~\cite{saade2018spoken}. This model is not quantized and has about 2.6M parameters.

\section{Keyword Spotting Method}
\label{sec:asr-lite}

We build an ASR-based keyword spotting method, similar to other existing CTC-based approaches~\cite{hwang2015online,lengerich2016end,chen2018sequence}.
The goal is to search for the keyword phone sequence in the predictions of the acoustic model. We compute a confidence score for every keyword in all segments of the prediction sequence, and then search for the best keyword sequence. We present the keyword detection method in Section~\ref{sec:asr-lite-detection}, the search for the keyword sequence in Section~\ref{sec:asr-lite-decoding}, the confidence scores explored in Section~\ref{sec:asr-lite-confidence} and some optimizations in Section~\ref{sec:asr-lite-optim}.

\subsection{Keyword detection}
\label{sec:asr-lite-detection}

In LVCSR, given an acoustic model and a language model, a search for the most likely sequence of words is carried out. In a keyword spotting approach, most of the words occurring in utterances are unknown, and the goal is to detect specific words or short phrases. Some methods based on LSTM and CTC are close to the LVCSR approach, where a filler model is inserted to replace all the \textit{unknown} words~\cite{hwang2015online,he2017streaming,chen2018sequence}. 

Here we adopt a different strategy. We consider all segments $[t_s, t_e]$, where $1 \leq t_s < t_e \leq T$ in the prediction sequence of length $T$. For every such segment, for each keyword $k$ we compute a confidence score $C(k, t_s, t_e)$ (cf. Section~\ref{sec:asr-lite-confidence}). From these, we build a set of detection candidates for a threshold $\tau$:
\begin{equation}
    \mathcal{C}_\tau = \{(k, t_s, t_e)\ : C(k, t_s, t_e) > \tau \}.
\end{equation}

We implemented a trie-based decoding, slightly better in complexity than scoring all segments and keywords separately. The set of keywords is converted to a prefix trie of the pronunciations. The decoding is implemented as a token passing algorithm. At each time step $t$, a new token is inserted at the root of the trie. All existing tokens are propagated based on the predictions of the network at $t$. A new candidate is created for each token in the terminal nodes if the confidence score exceeds the threshold $\tau$. We discuss how we improve the complexity in Section~\ref{sec:asr-lite-optim}.

\subsection{Finding the best keyword sequence}
\label{sec:asr-lite-decoding}


The goal of the post-processing is to build the final detection list $\mathbf{D}_\tau = (k_1, t_{s,1}, t_{e,1}) \ldots (k_n, t_{s,n}, t_{e,n})$, from elements of $\mathcal{C}_\tau$ such that $t_{e, i-1} < t_{s,i}$ for all $i$, i.e. the detected keywords are not overlapping. This could help in ambiguous situations.
For example, if the set of keywords is ``\texttt{play, playlist, stop}'', the query ``play the playlist top fifty'' could be ambiguous, and we might have overlapping detections of all three keywords in the ``playlist top'' segment.

We explored two strategies to obtain that sequence: a greedy approach where the keyword is output as soon as it is detected, and a full search for the best sequence that considers all possible non-overlapping detection sequences.

\subsubsection{Greedy}
In the \textit{greedy} approach, we look at the candidate detections $(k, t_s, t_e)$ in increasing order of end time $t_e$. When we find one for a given $t_e$, we add it to the list $\mathbf{D}_\tau$ (if we find several we keep the one with the highest confidence score) and remove from $ \mathcal{C}_\tau$ all candidates that overlap (cf. Algorithm~\ref{algo:greedy}).

\begin{algorithm}
\caption{Greedy decoding\label{algo:greedy}}
\begin{algorithmic}
\REQUIRE $\mathcal{C}_\tau$: detection candidates
\STATE $\mathbf{D}_\tau \leftarrow \emptyset$
\FOR{$t=1..T$}
\STATE $c_{t,\tau} = \{(k, t_s, t_e) \in \mathcal{C}_\tau : t_e = t \}$
\IF{$c_{t,\tau} \neq \emptyset$}
\STATE{$\mathbf{D}_\tau \leftarrow \mathbf{D}_\tau \cup \{\arg\max_{(k, t_s, t_e) \in c_{t,\tau}} C(k, t_s, t_e)\}$}
\STATE{$\mathcal{C}_\tau \leftarrow \mathcal{C}_\tau \setminus \{(k', t_s', t_e') \in \mathcal{C}_\tau : t_e > t_s'\}$}
\ENDIF
\ENDFOR
\RETURN $\mathbf{D}_\tau$
\end{algorithmic}
\end{algorithm}


\subsubsection{Sequence} 
The greedy approach does not guarantee that the best sequence of keywords will be output. For instance, it is likely that \texttt{play} will be detected with the greedy approach for ``launch my playlist'' even if \texttt{playlist} is in the keywords set. In the \textit{sequence} approach, all the candidates are considered and the sequence of non-overlapping keyword with the maximum cumulative confidence is selected:
\begin{equation}
    \mathbf{D}_\tau  = \arg\max_{NO(\mathcal{C}_\tau)} \sum_i C(k_i, t_{s,i}, t_{e,i}),
\end{equation}
where $NO(\mathcal{C}_\tau)$ is the set of sequences of non-overlapping elements of $\mathcal{C}_\tau$:
\begin{eqnarray*}
    NO(\mathcal{C}_\tau) & = & \{ (k_1, t_{s,1}, t_{e,1}) \ldots (k_n, t_{s,n}, t_{e,n}) \\
    &  & : \forall i, (k_i, t_{s,i}, t_{e,i}) \in \mathcal{C}_\tau ; t_{s_i} > t_{e,i-1}\}.
\end{eqnarray*}

\subsection{Confidence score computation}
\label{sec:asr-lite-confidence}

\subsubsection{CTC-based confidence score}

The CTC framework readily provides a method to compute the probability of a label sequence. For a segment $[t_s; t_e]$, the probability of keyword $k$ is given by:

\begin{equation}
    p(k | \mathbf{x}, t_s, t_e) = \sum_{\labseq : \mathcal{B}(\labseq)=k} \prod_{t=t_s}^{t_e} p(l_t | \mathbf{x}).
\end{equation}

With the Viterbi approximation, we can define the \textit{raw} confidence:

\begin{equation}
    C_{raw}(k, t_s, t_e) = \max_{\labseq : \mathcal{B}(\labseq)=k} \prod_{t=t_s}^{t_e} p(l_t | \mathbf{x}).
\end{equation}

The main problem with that computation arises from the fact that the networks make local predictions. The number of factors in the multiplication of local probabilities is equal to the length of the segment. So even if the network assigns a probability of $0.99$ in all the frames to the phones of the correct keyword, the resulting confidence will be $0.90$ for a segment of $10$ frames, and $0.74$ for a segment of $30$ frames. Therefore, the confidence will tend to be smaller for longer keywords and would not reflect the confidence of the network predictions. In the following, we discuss how we set a more meaningful confidence score.

\subsubsection{Segment length normalization}

One way to make the confidence score independent from the length of the segment is to normalize it by the segment length~\cite{bernardis1998improving}. We perform that normalization in the log space to compute $C_{nf}$:
\begin{equation}
    C_{nf}(k, t_s, t_e) = \exp \left( \frac{\log C_{raw}(k, t_s, t_e)}{t_e - t_s} \right),
\end{equation}
which amounts to take the exponential of the average frame log-likelihood of the segment as confidence score. 

\subsubsection{No-blank normalization}
CTC-trained networks tend to mostly predict \textit{blanks} with high probability and the labels quite locally. At first approximation:
\begin{eqnarray}
    p(\mathbf{l} | x_{t_s:t_e}) & = & \prod_{t=t_s}^{t_e} \mathbb{1}_{l_t = \oslash} p(\oslash | \mathbf{x}) \prod_{t=t_s}^{t_e} \mathbb{1}_{l_t \neq \oslash} p(l_t | \mathbf{x})\\
     & \approx & 1^{n_b} \times \prod_{t=t_s}^{t_e} \mathbb{1}_{l_t \neq \oslash} p(l_t | \mathbf{x})\\
     & \approx & \prod_{t=t_s}^{t_e} \mathbb{1}_{l_t \neq \oslash} p(l_t | \mathbf{x}), \label{eq:noblank_approx}
\end{eqnarray}
where $n_b$ is the number of blanks in $\mathbf{l}$. With this approximation, we see that the segment length normalization will tend to favor longer segments with more blanks, reducing the impact of label predictions. Since blanks are neither informative nor discriminating, we would like to normalize the score by the number of meaningful frames in the segments, i.e. the number of factors in the product in Eq.~\ref{eq:noblank_approx}, that is $t_e-t_s-n_b$. We can approximate
\begin{equation}
    n_b \approx \sum_{t=t_s}^{t_e} p(l_t = blank | \mathbf{x})
\end{equation}
and define the \textit{``noblank''} confidence score $C_{nb}$ as:
\begin{equation}
    \log C_{nb}(k, t_s, t_e) = \frac{\log C_{raw}(k, t_s, t_e)}{\sum_{t=t_s}^{t_e} (1 - p(l_t = \oslash | \mathbf{x}))}.
\end{equation}

\subsubsection{Likelihood ratio}
It is common in confidence score estimation to compute a likelihood ratio~\cite{weintraub1995lvcsr}. 
For example, with generative models:
\begin{equation}
    C(k) = \frac{\log p(\mathbf{x} | k)}{\log p_{background}(\mathbf{x})}
\end{equation}
where $p_{background}$ is computed with a background model.

In the same vein, we may calibrate the confidence score by computing the ratio between the keyword probability and the probability of the best label sequence given the outputs of the network~\cite{hwang2015online}. The probability of the best label sequence $C^*$ is given by
\begin{eqnarray*}
    C^*_{raw}(t_s, t_e) & = & \max_{\labseq} \prod_{t=t_s}^{t_e} p(l_t | \mathbf{x}) \\
    & = & \prod_{t=t_s}^{t_e} \max_{l_t} p(l_t | \mathbf{x})
\end{eqnarray*}
and we compute the normalized confidence as:
\begin{equation}
    C^{(r)}_{raw}(k, t_s, t_e) = \frac{C_{raw}(k, t_s, t_e)}{C^*_{raw}(t_s, t_e)}.
\end{equation}
This confidence equals one when the best label sequence corresponds to the keyword, and is close to zero when it is very different. In general, it measures how close is the keyword prediction to the best label prediction.

\subsubsection{Normalization and ratio}
The ratio is still a product of positive factors smaller than one. Although they each will be higher than the initial probability, the obtained confidence will suffer the same issues regarding segment length and amount of blank frames. Thus we can also combine the normalization schemes with the ratio, for example with the segment length normalization:
\begin{eqnarray*}
    C^*_{nf}(t_s, t_e) & = & \exp \left( \frac{\log C^*_{raw}(t_s, t_e)}{t_e - t_s} \right) \\
    C^{(r)}_{nf}(k, t_s, t_e) & = & \frac{C_{nf}(k, t_s, t_e)}{C^*_{nf}(t_s, t_e)}.
\end{eqnarray*}

\begin{figure}[!t]
    \centering
    \includegraphics[width=\linewidth]{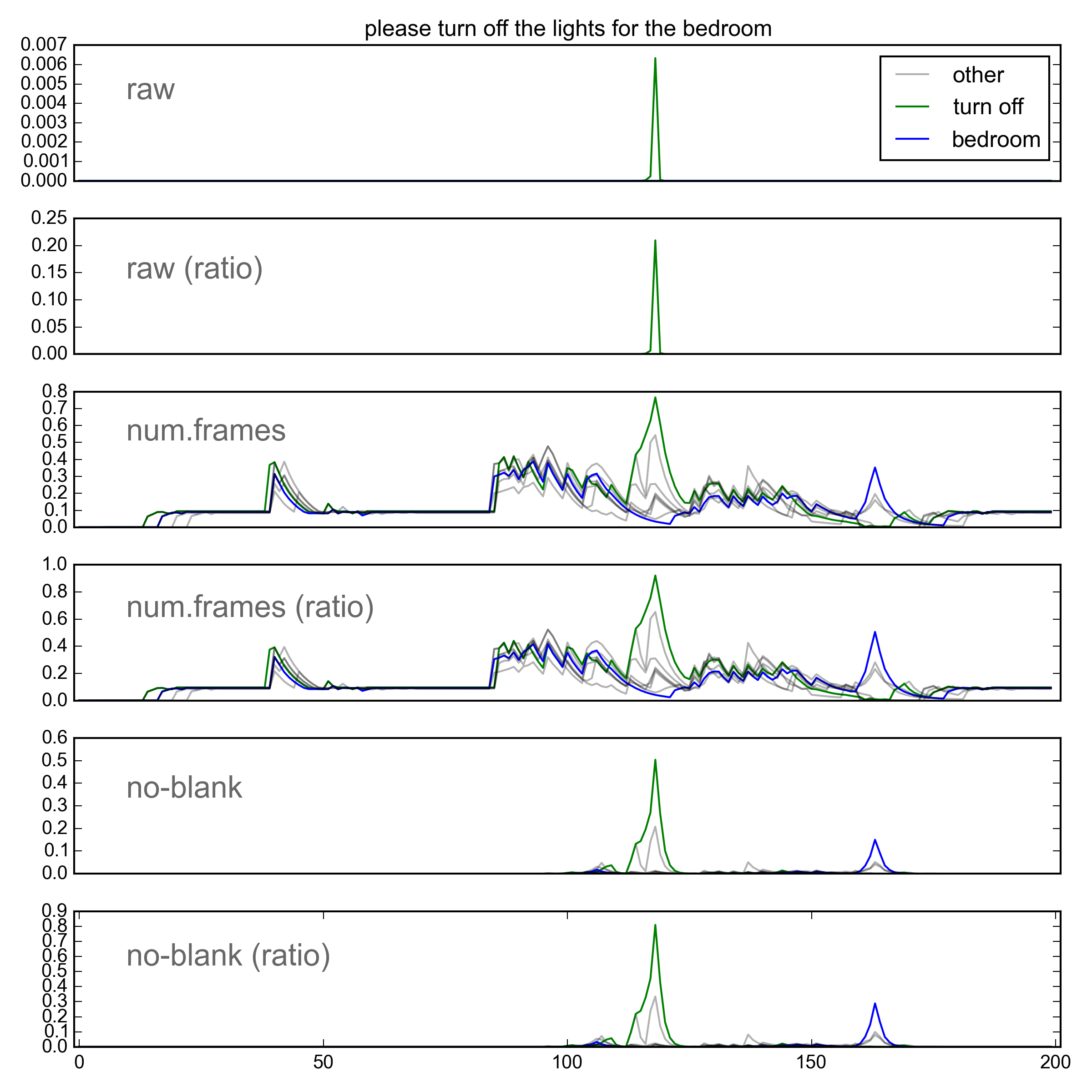}
    \caption{
    Comparison of confidence score calibration techniques.
    \label{fig:conf_score}}
    
\end{figure}

In Fig.~\ref{fig:conf_score}, we compare the different combinations of normalization and ratio for the query ``please turn off lights for the bedroom''. At each time step $t$ we plot the maximum confidence score of the two keywords appearing in the utterance ending at $t$. We observe that without any normalization, the confidence scores are quite low: 0.006 for \texttt{turn off}, \texttt{bedroom} is not detected. With the ratio, the confidence of \texttt{turn off} improves (around 0.2), but \texttt{bedroom} is still not detected. With the segment length normalization, the confidence scores are higher, and we see that it especially improved for \texttt{bedroom}. However, we also see many steps with relatively high confidence scores which do not correspond to keywords. With the ratio, the confidence scores are higher but follow the same trend. With the \textit{``no-blank''} technique, the confidence score of \texttt{bedroom} is lower but it looks like the most discriminative technique of all.

\subsection{Improving decoding speed}
\label{sec:asr-lite-optim}
Computing a confidence for all keywords in all segments is quite expensive, even with the trie implementation. The number of segments on which the computation of the confidence score should happen is $\mathcal{O}(T^2K)$ for a sequence of length $T$ and $K$ keywords. This will also impact the speed of the post-processing. We propose a few optimizations to improve the speed.

\subsubsection{Boundaries subsampling}
Since it is not crucial to find the exact boundaries of the segment, we can consider starting times only every three frames for example (i.e. every $90$ms). This divides the number of segments explored by three. We can apply the same idea to ending times and add detections to $\mathcal{C}_\tau$ only every three frames. This has no impact on the number of computation in the trie, since the score for segments $[t_s, t_e - 2]$ and $[t_s, t_e - 1]$ will anyway be calculated during the computation of the score for $[t_s, t_e]$. However, it has an impact on the complexity of the post-processing, dividing by three the number of detection candidates to consider.

\subsubsection{Maximum segment length}
It is fair to assume that a given keyword will be uttered in a limited amount of time. It should not be necessary to consider too long segments. Usually, about one second is sufficient. We can reduce a lot the number of segments to score by only computing scores for segments $[t_s; t_e]$ shorter than some predefined duration $S_{max}$ ($t_e - t_s < S_{max}$).

\subsubsection{Pruning}
The keyword detection scores are computed iteratively in a prefix trie. If the prefix of a keyword has a low probability, the whole keyword is likely to have a low probability. 
In the token passing algorithm, we drop any path for which the average negative log-likelihood per frame is higher than $2.5$. 

\begin{figure}[!t]
    \centering
    \includegraphics[width=\linewidth]{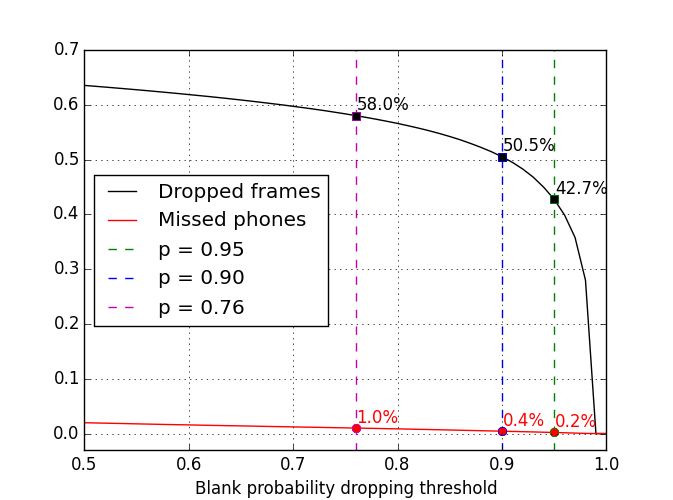}
    \caption{Amount of dropped frames, and missed phones for different thresholds on the blank probability, measured on an aligned dataset.\label{fig:hspike}}
\end{figure}

\subsubsection{Ignoring blank frames}
We have seen that the blank predictions could be problematic to compute the confidence score because they are not informative and dominate the prediction sequence. Taking inspiration from the phone-synchronous decoding \cite{chen2016phone,zhuang2016unrestricted}, we drop the prediction frames when the probability of the blank label exceeds some threshold. It amounts to realizing the approximation of Eq.~\ref{eq:noblank_approx}. To measure the impact of this strategy, we align a dataset and compute the ratio of dropped frames and missed phones at different dropping thresholds. The results are displayed in Fig.~\ref{fig:hspike}. We observe that we can almost drop 60\% of the frames while missing less than 1\% of actual phones. That would represent 60\% less computation in each segment.

\subsection{Online keyword spotting}

Since the acoustic model is only made of dense layers and unidirectional LSTM, we can use it in a streaming, online mode, where we feed the MFCC frames as the audio comes, and output a new prediction frame every 30ms. In the keyword detector, the tokens are updated at every prediction frame, and candidate detections ending at this time step are produced. This is therefore also compatible with a streaming mode. The greedy post-processor outputs detections as soon as the exceed the threshold, so it is easily applicable to online scenario. The sequence post-processor should remember all best sequences ending anywhere between $t$ and $t - S_{max}$ (where $S_{max}$ is the maximum segment length), and must wait for the end of the query to output the final detected sequence of keywords. However, the computation itself can be done in streaming.


\section{Experimental setup}
\label{sec:setup}
In this section, we present the experimental setup: the training data and procedure, the evaluation tasks and associated datasets, the metrics with which we evaluate our system, and the baselines we compare it to.

\subsection{Training}
We train the acoustic model on the Librispeech dataset~\cite{panayotov2015librispeech}. The training set contains 960 hours of English read speech. To make the model robust to noisy far-field environments, we augment the training data four times. We use the \texttt{pyroomacoustics} library~\cite{scheibler2018pyroomacoustics} to simulate random rooms and speaker and microphone positions, with random noise sources. We train the networks with CTC~\cite{graves2006connectionist} to predict phone sequences. 

The non-quantized LSTM networks are first trained for $40$ epochs on the augmented Librispeech training set.
After this first step, they are converted to use the quantized LSTM cell, and further trained for five epochs,
with the same training hyperparameters. 
A pronunciation model combining a flat lexicon with a grapheme-to-phoneme converter is used to convert the 
transcripts of the dataset to phone sequences that serve as target for the training.

\subsection{Evaluation tasks and datasets}
We evaluate our system in a mini-SLU scenario.
In this scenario, we suppose that the system has been triggered by the user, for example by saying a wake word.
The goal is to detect the keywords of interest from the subsequent query.
We defined two tasks corresponding to a smart lights scenario and a washing machine scenario. For each task, we selected eight keywords:
\texttt{turn on, turn off, increase, decrease, brightness, kitchen, living room, bedroom} for smart lights, 
\texttt{hot water, cold water, high spin, low spin, wash heavy duty, wash normal, wash colors, wash delicate} for washing machine.

\begin{table}[!t]
\centering
\begin{tabular}{lcc}
        & lights & washing \\\hline
Samples & 564 & 545 \\
Unique keywords & 8 & 8 \\
Speakers (M/F) & 32 (22/10) & 33 (22/11) \\
Samples/speaker -  avg (min/max) & 18 (8/60) & 17 (5/50) \\
Duration (s) - avg (min/max) & 2.6 (1.6/6.1) & 3.4 (1.8/6.7) \\\hline
\end{tabular}   
\caption{Mini-SLU datasets statistics.\label{tab:data}}
\end{table}

We crowd-sourced over five hundred queries for each use-case from over $30$ speakers. Each query contains between one and four keywords, and are expressed in natural language (e.g. ``could you \texttt{[turn on]} the lights in the \texttt{[bedroom]}"). Each dataset was re-recorded in clean and noisy, reverberated far-field conditions with a SNR of 5dB. We present some statistics of these datasets in Table~\ref{tab:data}. 
The datasets will be made publicly available.

\subsection{Evaluation metrics}

We evaluate our models with two metrics. At the keyword level, we measure the F1 score, which illustrate the ability of the system to pick up all and only the keywords. For a simple keyword-based SLU system, it is important that the whole query is correctly parsed, i.e. that the correct sequence of keywords is detected. We measure this by computing the ratio of exactly parsed queries, i.e.~those for which the sequence of detected keywords exactly matches the expected one.

\subsection{LVCSR-based KWS baselines}

The first baselines are based on the decoding of the queries with a large vocabulary and a language model, similar to the ones used for ASR tasks. Two such baselines are evaluated with both the large TDNN-LSTM network and the best quantized LSTM network. The first one consists in looking for the keywords in the Viterbi decoding of the query. For the second one, we first extract recognition lattices and compute word posteriors in the lattice. The keywords are then searched for in the lattice.

\subsection{Filler model baselines}

For the filler model baseline, a decoding graph is built with two parallel paths: one for the keywords and one for a phone-loop filler model. The false alarm rate is controlled by adjusting the transition cost from the filler model to the keywords. The output of this baseline is derived from a Viterbi decoding in this graph. We evaluated this baseline for the large TDNN-LSTM network and the best quantized LSTM network.

\subsection{CTC-KWS baselines}

Finally, two methods for CTC-based KWS were implemented: a phone-synchronous decoding mimimum edit distance~(PSD-MED) approach derived from~\cite{zhuang2016unrestricted,chen2018sequence} and a CTC-decoding one similar to~\cite{hwang2015online}.

The PSD-MED baseline implements the decoding method of Zhuang et al.~\cite{zhuang2016unrestricted,chen2018sequence} on top of the best quantized LSTM neural network trained for this paper. The main differences with the reference papers are that there is no word boundary class for that neural network, that the confidence scores are normalized by the number of frames, and that the ``sequence'' post-processing is applied to the output of the method. The results we obtained, in particular how they compare to keyword-filler baseline, are consistent with the results reported in~\cite{chenthesis}.

For the CTC-decoding baseline, we set up our system to match the approach proposed by Hwang et al.~\cite{hwang2015online} as closely as possible. The confidence scores are normalized by the length of the detected segments and the decision is made based on the score ratio, with the greedy approach. The main difference with the reference paper is the absence of a word boundary class.

\section{Results}
\label{sec:results}
In this part we present the detailed results of the experiments for the proposed approach. The first step is to train a generic ASR acoustic model. We give in Section~\ref{sec:res-am-lvcsr} the error rates of the different trained models in an LVCSR setup. The purpose is to give an idea of the performance of the obtained models, to put the subsequent results in perspective. 
In Section~\ref{sec:res-conf-score}, we compare the results obtained with different confidence score strategies. The post-processing methods are compared in Section~\ref{sec:res-post-proc}. The impact of the presented techniques to improve the decoding speed is measured in Section~\ref{sec:res-speed}. We show the impact of model size in Section~\ref{sec:res-size} and of quantization in Section~\ref{sec:res-quant}. Finally, we compare the proposed approach to the baselines in Section~\ref{sec:res-compare}.

\subsection{Acoustic model training}
\label{sec:res-am-lvcsr}

We train the base acoustic models according to the method presented in Section~\ref{sec:am-training}. 
The models are stacks of 3 or 5 unidirectional LSTM layers, with 64, 96 or 128 units in each layer.
They are trained to minimize the CTC loss~\cite{graves2006connectionist} with the Adam optimizer with 
minibatches of 32 samples and a learning rate of $ 1e-3$, annealed by a factor $0.9$ every time that 
no improvement has been seen for $3000$ updates. We applied a curriculum strategy to focus on the shorter 
samples at the beginning of training and add longer ones at each epoch. 

\begin{figure}[!t]
    \centering
    \includegraphics[width=\linewidth]{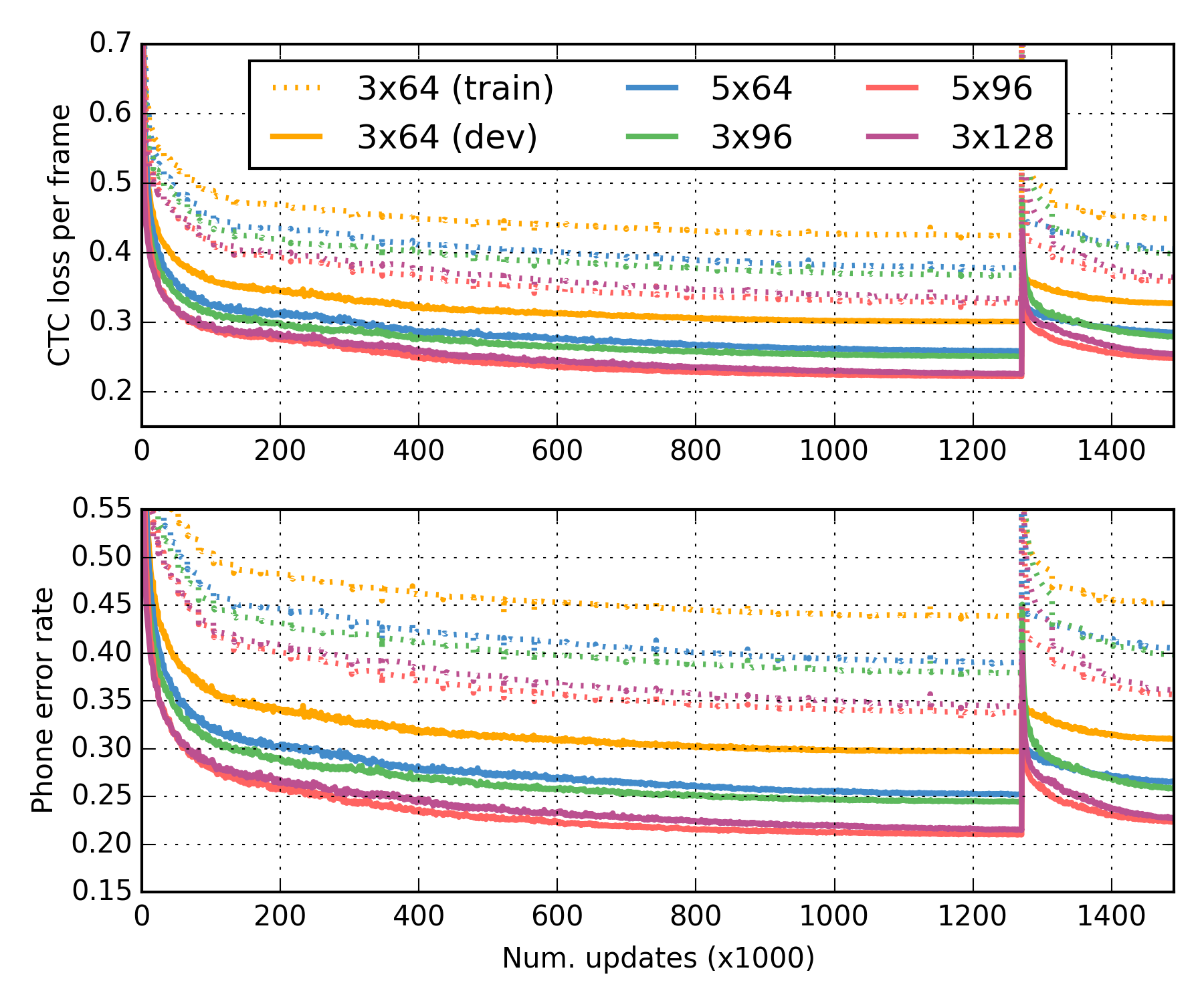}
    \caption{Convergence curves of CTC training of networks of different sizes. The top plot shows the CTC loss per frame. The bottom plot show the normalized edit distance between the raw CTC predictions and the ground-truth.}
    \label{fig:convergence-am}
\end{figure}

We plot the convergence curves for the five models we trained in Fig.~\ref{fig:convergence-am}.
As expected, bigger models yield lower error rates, and models of similar sizes yield similar error rates. When quantization is activated at epoch $40$, all networks suffer a large loss of performance, almost entirely recovered after the quantized training. This loss of performance is mainly due to the quantization of the logits, which has a direct impact on the predictions of the network. 

Then, we plug the small neural network acoustic models into a large-vocabulary ASR setup. We use the standard vocabulary of 200k words provided with the Librispeech corpus\footnote{\url{http://www.openslr.org/11/}}. We applied a trigram language model pruned with threshold $3e{-7}$, also provided with Librispeech, sometimes referred to as \texttt{tgmed} in the literature. We carried out a single-pass Viterbi beam search, with a log-likelihood beam of 8. We measure the word error rate of the obtained ASR system on the development and test sets of Librispeech, and report the results in Table~\ref{tab:asr_libri}. We evaluate all the models, before quantization (i.e. at the end of the initial $40$ epochs of training) and after quantization (at the end of training).

\begin{table}[!t]
\begin{center}
    \begin{tabular}{rcccc}
     & \textbf{dev-clean} & \textbf{dev-other} & \textbf{test-clean} & \textbf{test-other} \\\hline
     3x64 & 23.0 & 44.5 & 22.8 & 47.1 \\
     (quantized) & 22.8 & 41.7 & 21.8 & 43.7 \\\hline
     5x64 & 17.5 & 36.8 & 17.2 & 39.1 \\
     (quantized) & 18.1 & 36.3 & 18.0 & 38.5 \\\hline
     3x96 & 16.4 & 35.7 & 16.7 & 37.7 \\
     (quantized) & 16.8 & 34.7 & 16.7 & 36.6 \\\hline
     5x96 & 13.7 & 30.7 & 13.7 & 31.9 \\
     (quantized) & 13.5 & 29.5 & 13.8 & 31.0 \\\hline
    3x128 & 13.8 & 31.3 & 14.2 & 33.0 \\
     (quantized) & 13.8 & 29.9 & 13.9 & 30.9 \\\hline
     TDNN-LSTM & 7.3 & 16.6 & 7.4 & 17.4 \\\hline
\end{tabular}
\end{center}
    \caption{Word Error Rates (\%) obtained with different acoustic models on LibriSpeech, with the \texttt{tgmed} language model.}
    \label{tab:asr_libri}
\end{table}

Again, we logically get better results with bigger models. The error rates achieved with the quantized models tend to be slightly lower than the corresponding results of the non-quantized ones. Although it may seem counter-intuitive, it is possible that this effect is merely due to the additional five epochs of training. Finally, we note that these results are quite far from the current state-of-the-art for this dataset. However, the models we present are very small, and the error rates result from a single-pass streaming decoding. They are given as an indication of the performance of the trained network as an acoustic model, and look acceptable from this perspective.

We applied the same LVCSR decoding to the proposed dataset and measured the word error rates for the quantized 5x96 LSTM network and for the large TDNN-LSTM model. 
The results are reported in Table~\ref{tab:asr_slu}. We see that these datasets are quite challenging. Even the TDNN-LSTM yields high error rates, more than five times those obtained on Librispeech.

\begin{table}[!t]
\begin{center}
    \begin{tabular}{rcccc}
    & \multicolumn{2}{c}{\textbf{lights}} & \multicolumn{2}{c}{\textbf{washing}} \\
     & clean & noisy & clean & noisy\\\hline
     5x96 (quantized) & 54.6 & 79.3 & 72.8 & 87.9 \\
            TDNN-LSTM & 32.8 & 59.6 & 46.5 & 68.6 \\\hline
\end{tabular}
\end{center}
    \caption{Word Error Rates (\%) obtained with different acoustic models on the mini-SLU datasets, with the \texttt{tgmed} language model.}
    \label{tab:asr_slu}
\end{table}



\subsection{Confidence score calibration}
\label{sec:res-conf-score}

\begin{figure}[!t]
    \centering
    \includegraphics[width=\linewidth]{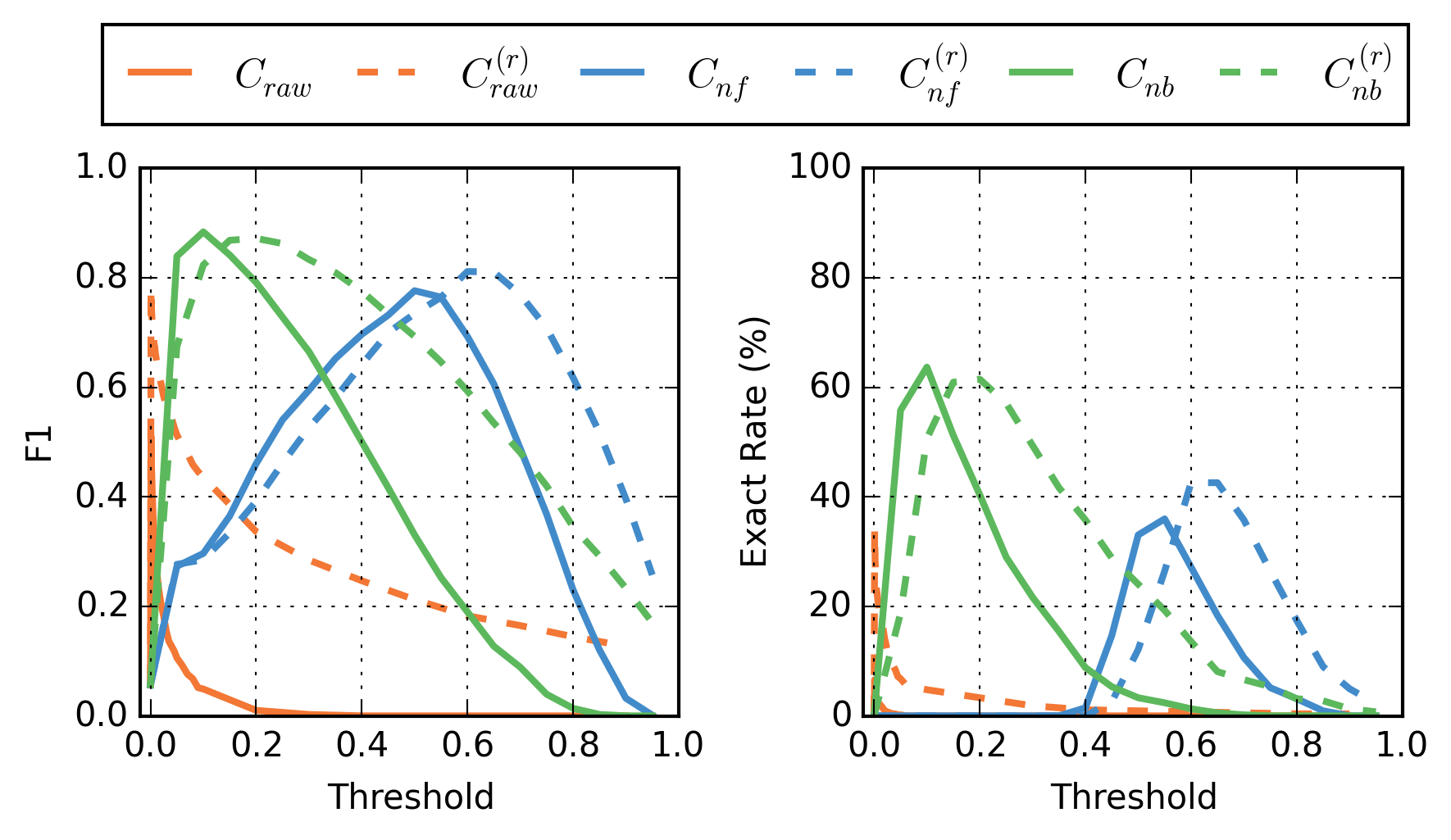}
    \caption{Comparison of F1-scores and exact rates for different \textbf{confidence score calibration} methods on the \texttt{washing-clean} dataset using the not quantized \texttt{3x96} network.}
    \label{fig:conf-score-comp}
\end{figure}

We have seen in Section~\ref{sec:asr-lite} that the definition of a good confidence score was quite important to aggregate the framewise phone-level scores of the acoustic model into a meaningful keyword-level confidence. We first compare the different strategies for normalization ($C_{raw}$, $C_{nf}$ and $C_{nb}$), with or without ratio ($C_{*}$, $C^{(r)}_{*}$). We plot the F1 score and exact rate results in Fig.~\ref{fig:conf-score-comp}, with the not quantized \texttt{3x96} network on the \texttt{washing-clean} dataset.

We see that the different normalization strategies are associated with different optimal thresholds. For reasons we already explained, without any normalization, most of the confidence scores are quite low. As expected, normalizing by the number of frames ($C_{nf}$) gives more understandable results. Indeed, the confidence in that case can be interpreted as a per-frame probability. The ``\textit{no blank}'' normalization yields the best results. It tends to give lower scores in general, but scores much closer to zero for incorrect detection candidates, reducing the risk of false alarms.

Using score ratios as confidence (dashed lines) tends to yield higher scores and optimal thresholds. Moreover, the performance seems to decrease more smoothly when the threshold deviates from the optimal one, giving a bit more robustness to the system. It also improves the performance for segment length normalization or when there is no normalization. For the ``\textit{no blank}'' normalization, the performance is slightly lower.

\subsection{Post-processor}
\label{sec:res-post-proc}

The previous results were obtained with the greedy post-processor. We now compare the performance of the greedy post-processor with the sequence one which provides a more elaborated enforcement of non-overlapping constraints.  

\begin{figure}[!t]
    \centering
    \includegraphics[width=\linewidth]{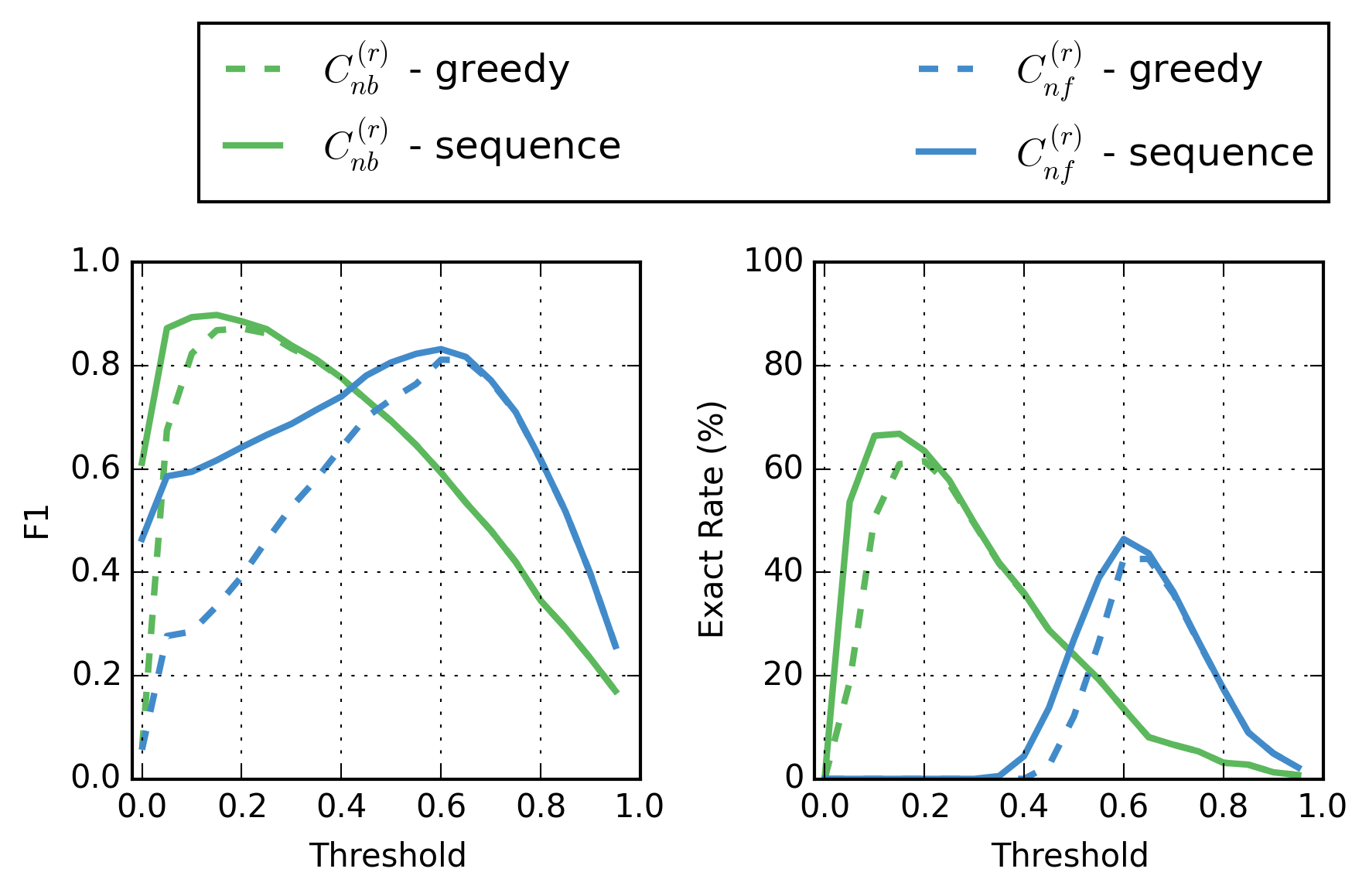}
    \caption{Comparison of F1-scores and exact rates for \textbf{different post-processors} on the \texttt{washing-clean} dataset using the not quantized \texttt{3x96} network.}
    \label{fig:asr-lite-pp}
\end{figure}

The results are displayed in Fig.~\ref{fig:asr-lite-pp} for the confidence scores using the score ratio and the normalizations by the number of frames and or non-blank frames ($C^{(r)}_{nf}$ and $C^{(r)}_{nb}$). 
With higher threshold, the false rejection rate increases. That is not recovered by using a different post-processor. For lower threshold however, the greedy post-processor triggers more quickly, and the false alarm rate increases. Using the sequence post-processor, some of these false alarms are discarded by the non-overlap constraints. Therefore, there is a slower decrease of performance when the threshold decreases. For the \textit{``no-blank''} normalization ($C^{(r)}_{nb}$), we get better results than the greedy approach with a smaller threshold. 

\subsection{Decoding speed}
\label{sec:res-speed}

We presented in Section~\ref{sec:asr-lite} a few tricks to improve the decoding speed. They usually also result in less detection candidates. This should also increase the speed of the post-processor, but might result in a degraded detection accuracy. In the following, we will measure the processing time reduction as well as the performance degradation to find the best trade-off.

\begin{figure}[!t]
    \centering
    \includegraphics[width=\linewidth]{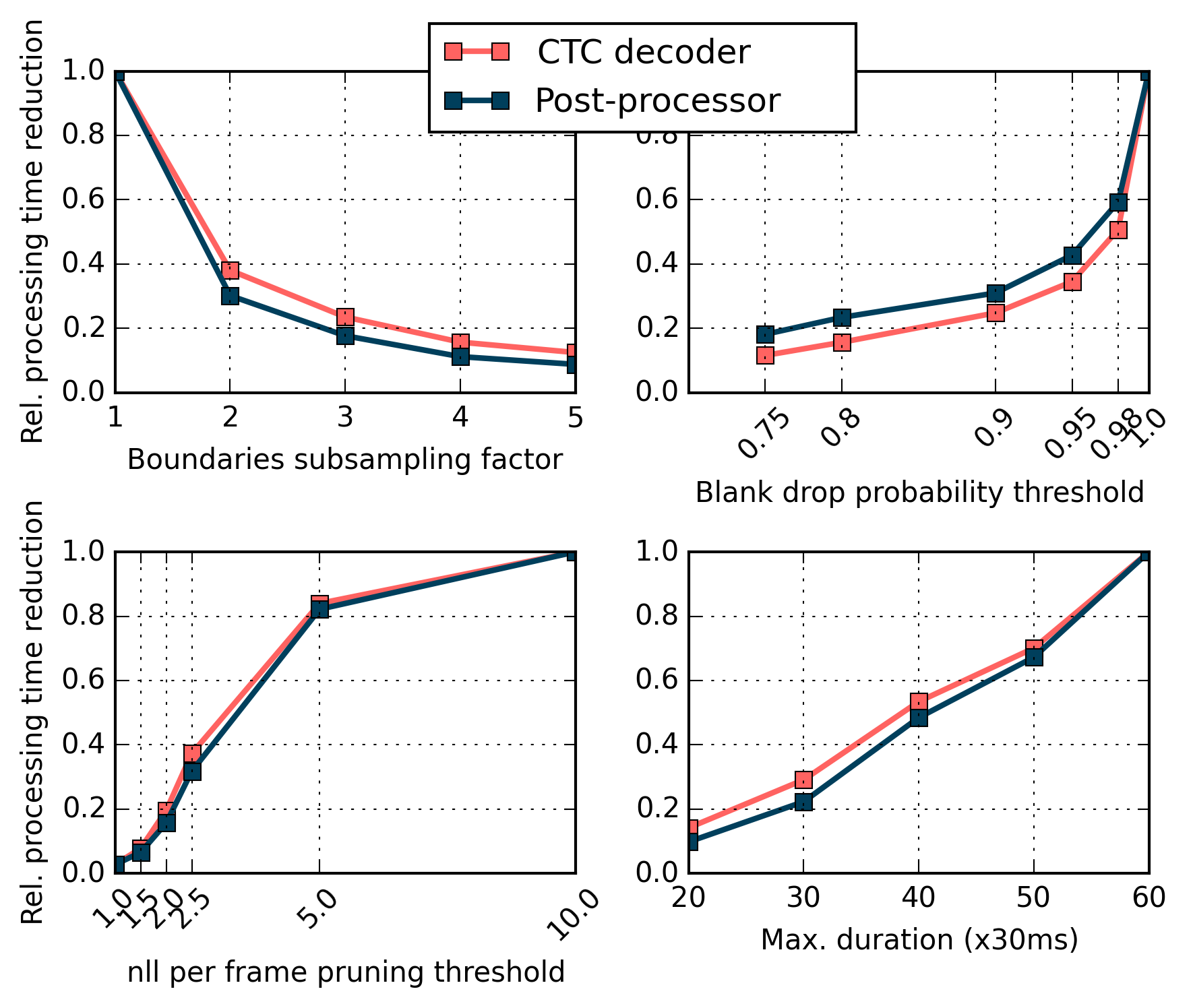}
    \caption{Processing time reduction on \texttt{washing-clean} dataset using the not quantized \texttt{3x96} network.}
    \label{fig:asr-lite-proc-times}
\end{figure}

In Fig.~\ref{fig:asr-lite-proc-times} we plot the relative processing time reduction for both the decoder and post-processor for the different tricks. We see that each of them can easily bring a improvement of processing time by a factor two for the decoder and the post-processor.

\begin{figure}[!t]
    \centering
    \includegraphics[width=\linewidth]{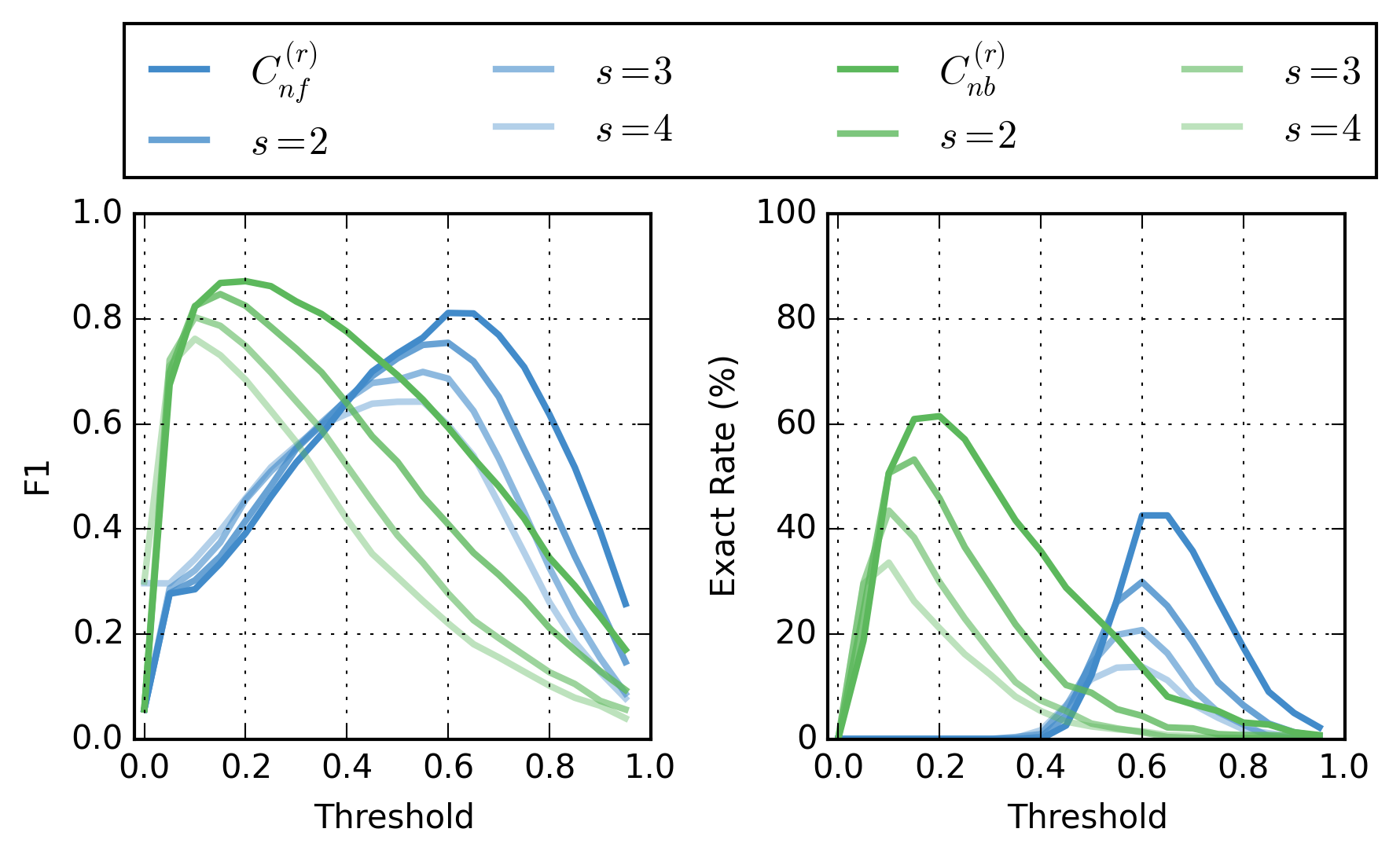}
    \caption{Comparison of F1-scores and exact rates for different \textbf{boundary subsampling factors} on the \texttt{washing-clean} dataset using the not quantized \texttt{3x96} network.}
    \label{fig:asr-lite-bound-ss}
\end{figure}

For the boundaries subsampling, we already achieve more than a factor two by considering only every other frame as possible boundary (top-left plot in Fig.~\ref{fig:asr-lite-proc-times}). However we notice in Fig.~\ref{fig:asr-lite-bound-ss} that the accuracy of the system decreases quickly with the subsampling factor, for two confidence score configurations.

\begin{figure}[!t]
    \centering
    \includegraphics[width=\linewidth]{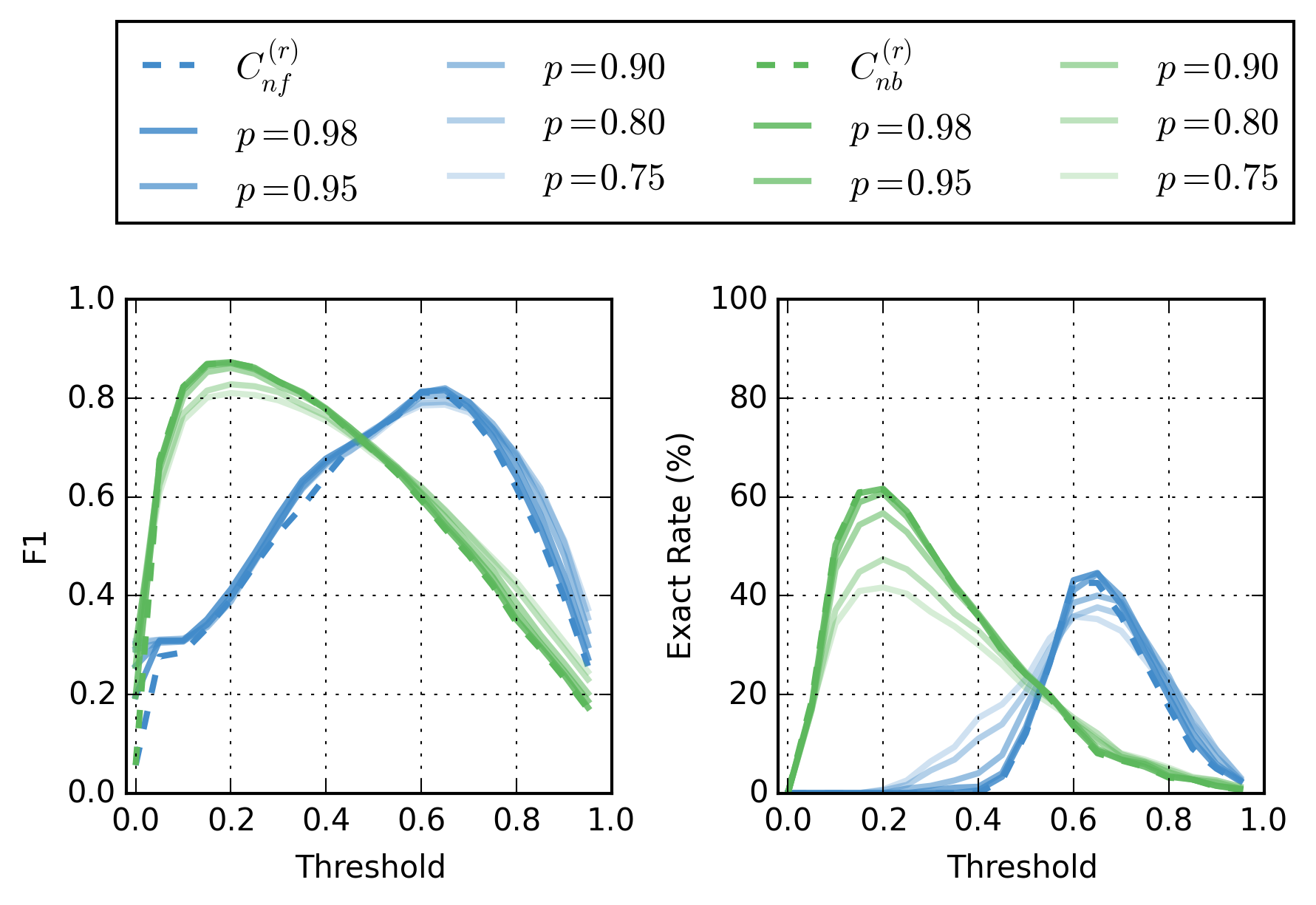}
    \caption{Comparison of F1-scores and exact rates for different \textbf{blank dropping probability thresholds} on the \texttt{washing-clean} dataset using the not quantized \texttt{3x96} network.}
    \label{fig:asr-lite-hspike-res}
\end{figure}

Another option was to drop completely in the search the frames where the blank probability exceeded some threshold. We saw in Fig.~\ref{fig:hspike} that at the expense of potentially dropping a few frames containing useful information, we could almost skip half the frames during decoding. The same acceleration factor is achieved with a quite high threshold of $0.95$ on the blank probability as with a boundary subsampling factor of 2 (top-right plot of Fig.~\ref{fig:asr-lite-proc-times}). We also see in Fig.~\ref{fig:asr-lite-hspike-res} that the system is more robust to that acceleration method than to the boundary subsampling one. With thresholds above $0.90$ we almost observe no degradation of performance.

\begin{figure}[!t]
    \centering
    \includegraphics[width=\linewidth]{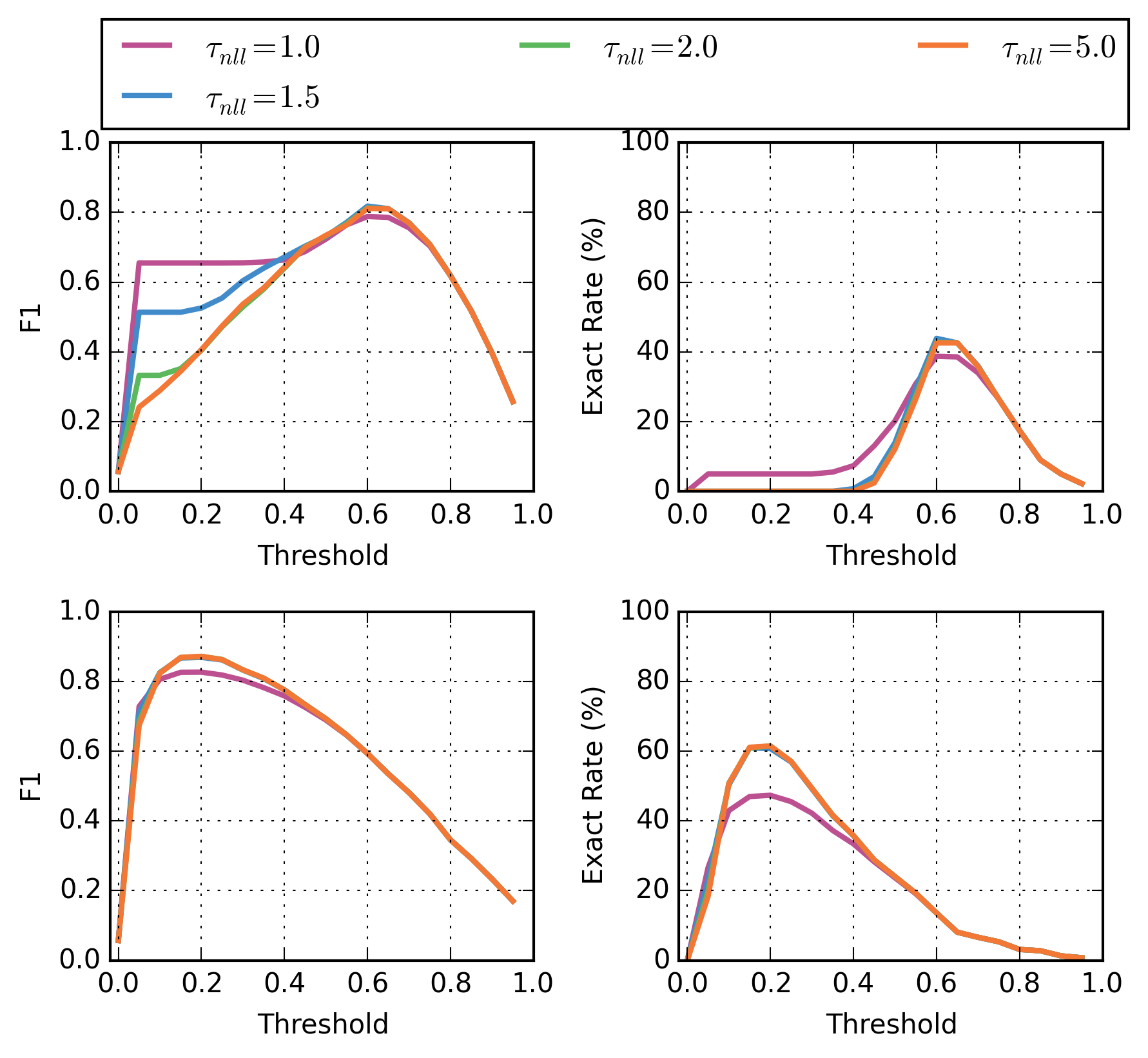}
    \caption{Comparison of F1-scores and exact rates for different \textbf{pruning thresholds} with the $C^{(r)}_{nf}$ (top) and $C^{(r)}_{nb}$ (bottom) confidence on the \texttt{washing-clean} dataset using the not quantized \texttt{3x96} network.}
    \label{fig:asr-lite-prune}
\end{figure}

Most of the evaluated segments will have a very different content from any of the keyword. That might be detected early in the decoding process. As a result, pruning strategies should allow to drop the segment early, and as we notice in the bottom-left plot of Fig.~\ref{fig:asr-lite-proc-times}, provide huge speed-ups. In Fig.~\ref{fig:asr-lite-prune}, we see that the performance remains almost the same for any pruning threshold above $1.5$. With the ``\textit{no-blank}'' confidence with ratio ($C^{(r)}_{nb}$, bottom), a threshold of $1.0$ induce a significant drop of exact rate. For the ``\textit{num. frames}'' confidence with ratio ($C^{(r)}_{nf}$), stricter threshold actually improves the performance for small threshold: it might be because it reduces the number of false alarms, which we have seen are numerous for this confidence.

\begin{figure}[!t]
    \centering
    \includegraphics[width=\linewidth]{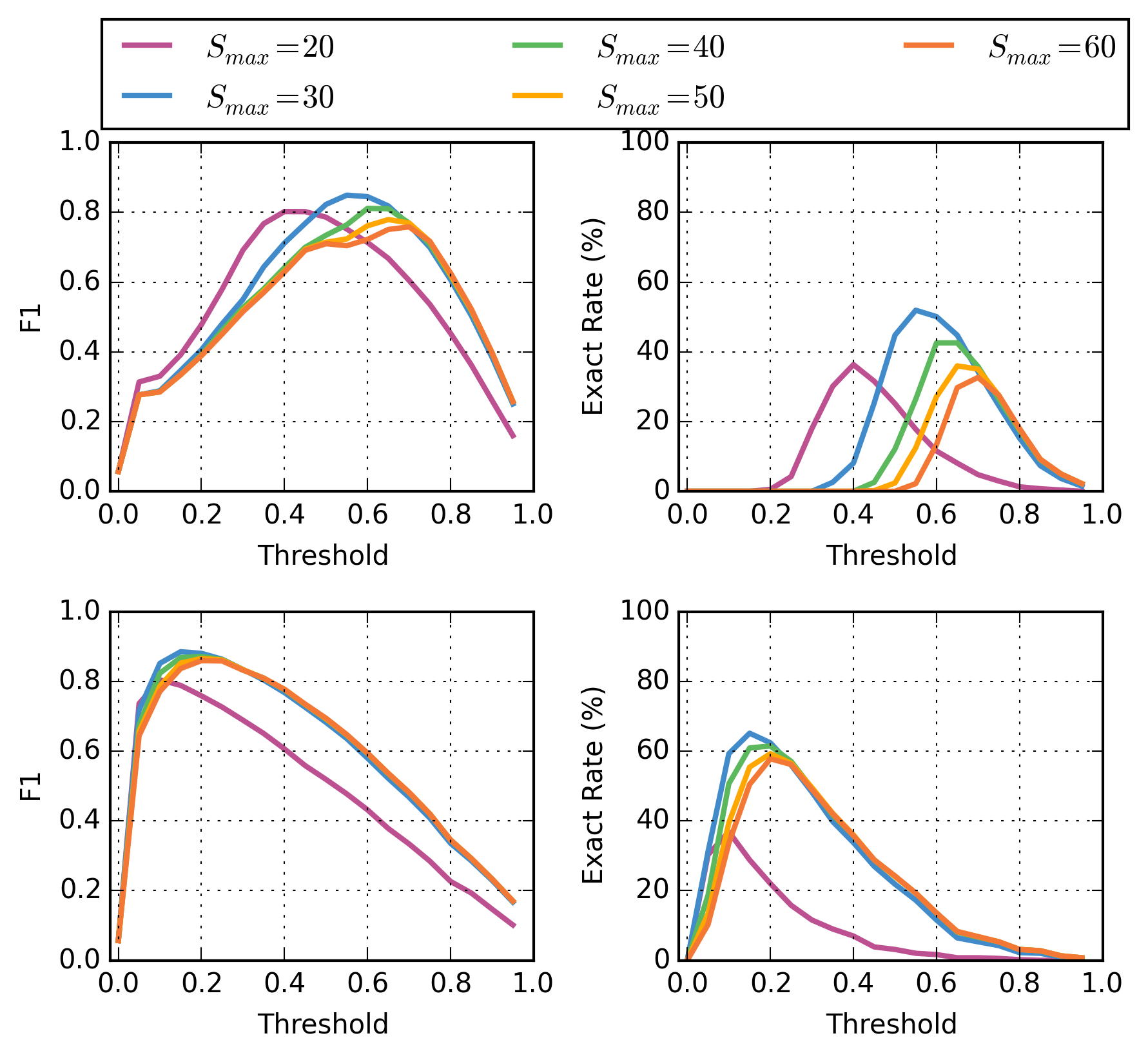}
    \caption{Comparison of F1-scores and exact rates for different \textbf{values of $S_{max}$} with the $C^{(r)}_{nf}$ (top) and $C^{(r)}_{nb}$ (bottom) confidence on the \texttt{washing-clean} dataset using the not quantized \texttt{3x96} network.}
    \label{fig:asr-lite-max-dur}
\end{figure}

Finally, the decoding speed decreases linearly with the maximum segment length (bottom-right plot of Fig.~\ref{fig:asr-lite-proc-times}). This parameter is closely related to the actual duration of a keyword. If it is set to a too small value, there will be a lot of false rejections of long keywords. If it is too big, the chances of unrealistic detections increases. This is verified in Fig.~\ref{fig:asr-lite-max-dur}, where we see that $S_{max}=20$ (i.e. 600ms) seems too small, while $S_{max}=30$ (900ms) appears to be the best choice. With bigger values, the performance decreases. This is especially true for the ``\textit{num. frames}'' confidence with ratio ($C^{(r)}_{nf}$): it is normalized by the segment length, so the impact of a single frame on the confidence is smaller for longer segments.

\subsection{Impact of acoustic model size}
\label{sec:res-size}

\begin{figure}[!t]
    \centering
    \includegraphics[width=\linewidth]{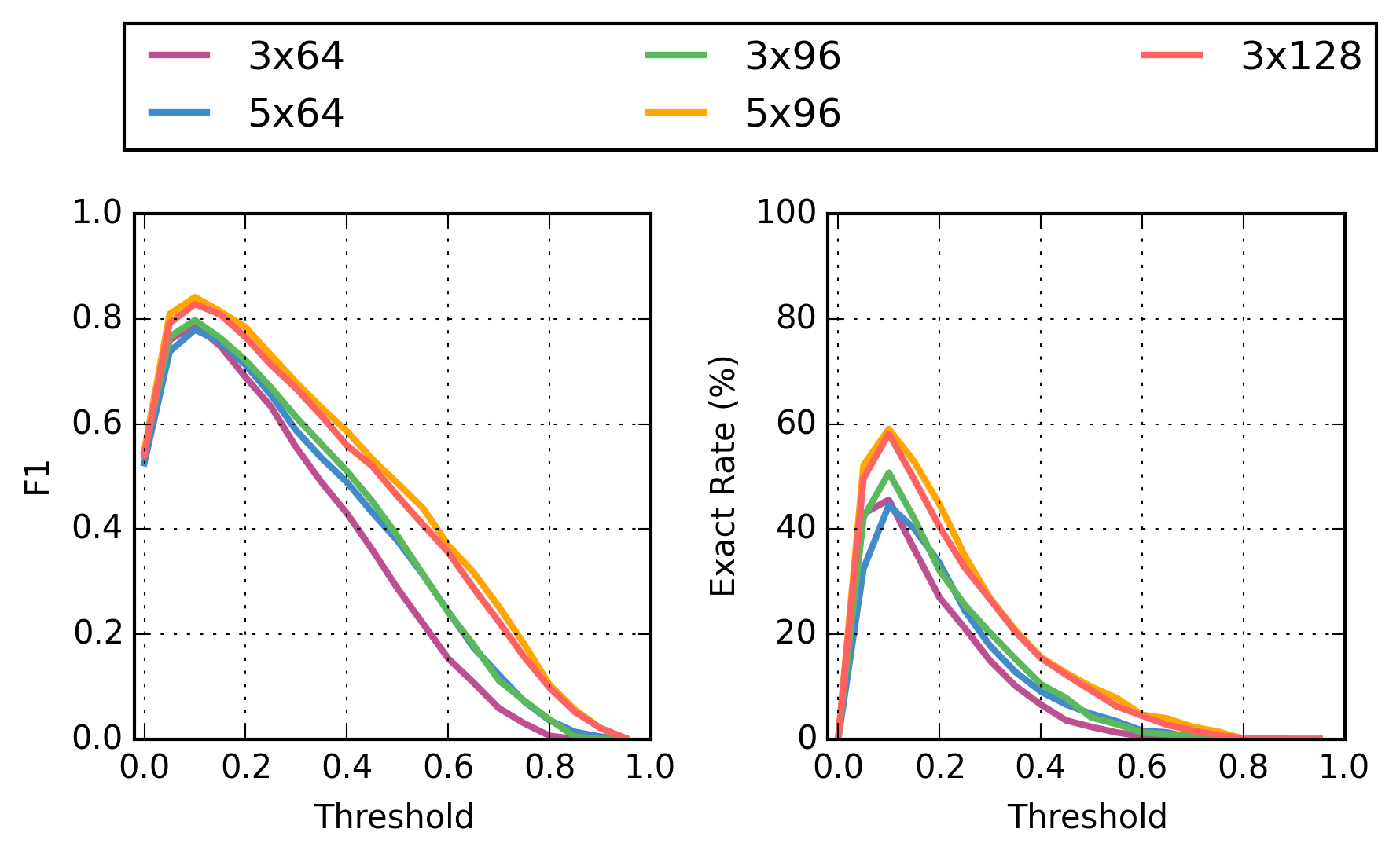}
    \caption{Comparison of F1-scores and exact rates for different \textbf{neural network sizes} on the \texttt{lights-clean} dataset (not quantized), with the $C_{nb}$ confidence.}
    \label{fig:am-size-impact}
\end{figure}

We now compare the results for acoustic models of different sizes to measure the impact of the number of parameters on the performance of the system. We plot the F1 score and exact rate results for all the trained models, before quantization on Fig.~\ref{fig:am-size-impact}. As expected, bigger models yield better results. Yet the difference between the smallest one (\texttt{3x64}) and the biggest (\texttt{3x128}) is not as big for the keyword detection task as it is for the LVCSR task of Section~\ref{sec:res-am-lvcsr}.

\subsection{Impact of quantization}
\label{sec:res-quant}

\begin{figure}[!t]
    \centering
    \includegraphics[width=\linewidth]{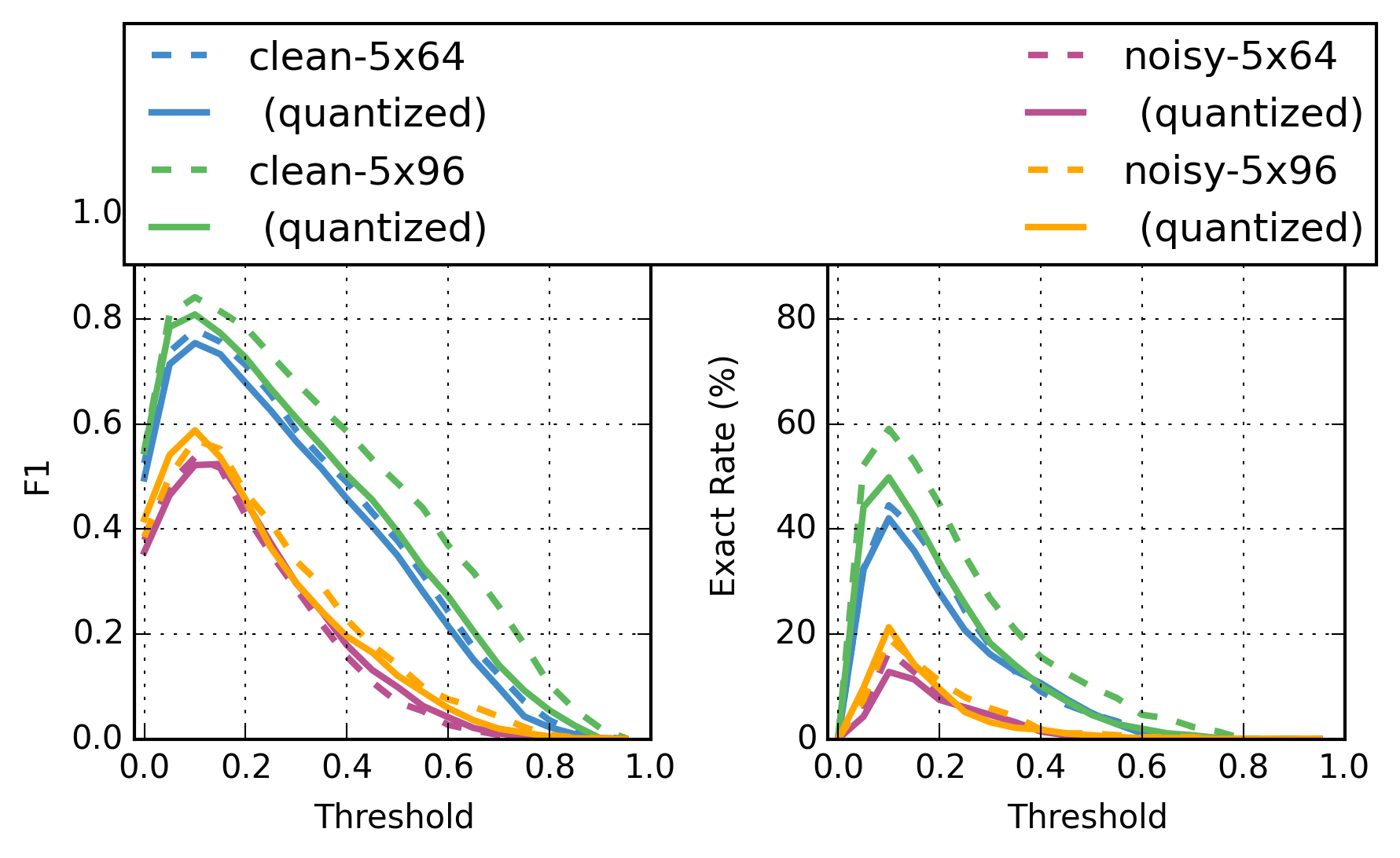}
    \caption{Comparison of F1-scores and exact rates for \textbf{quantized and not quantized networks} on the \texttt{lights} dataset (clean and noisy), with the $C_{nb}$ confidence.}
    \label{fig:quantization-impact}
\end{figure}

After a fixed number of epochs, the acoustic models are quantized and further trained. We have seen in Section~\ref{sec:res-am-lvcsr} that the quantized models, which have been trained longer, yield similar or better word error rates for LVCSR than the ones before quantization. We compare in Fig.~\ref{fig:quantization-impact} the effect of quantization on the performance of the systems for two model sizes and two datasets. We notice that quantized models tend to perform a bit worse than their floating-point counterparts. The difference in performance is not so big for noisy datasets but appears to be significant on clean ones.

\subsection{Final results}
\label{sec:res-compare}

To compare the systems and design choices (post-processor, confidence score, etc.), we select, for each case, the threshold that corresponds to the best cumulative exact rate across the four datasets. This threshold might not be the optimal one for an individual dataset but corresponds more closely to a real-world scenario where a single threshold is set for the system.

\begin{figure}[!t]
    \centering
    \includegraphics[width=\linewidth]{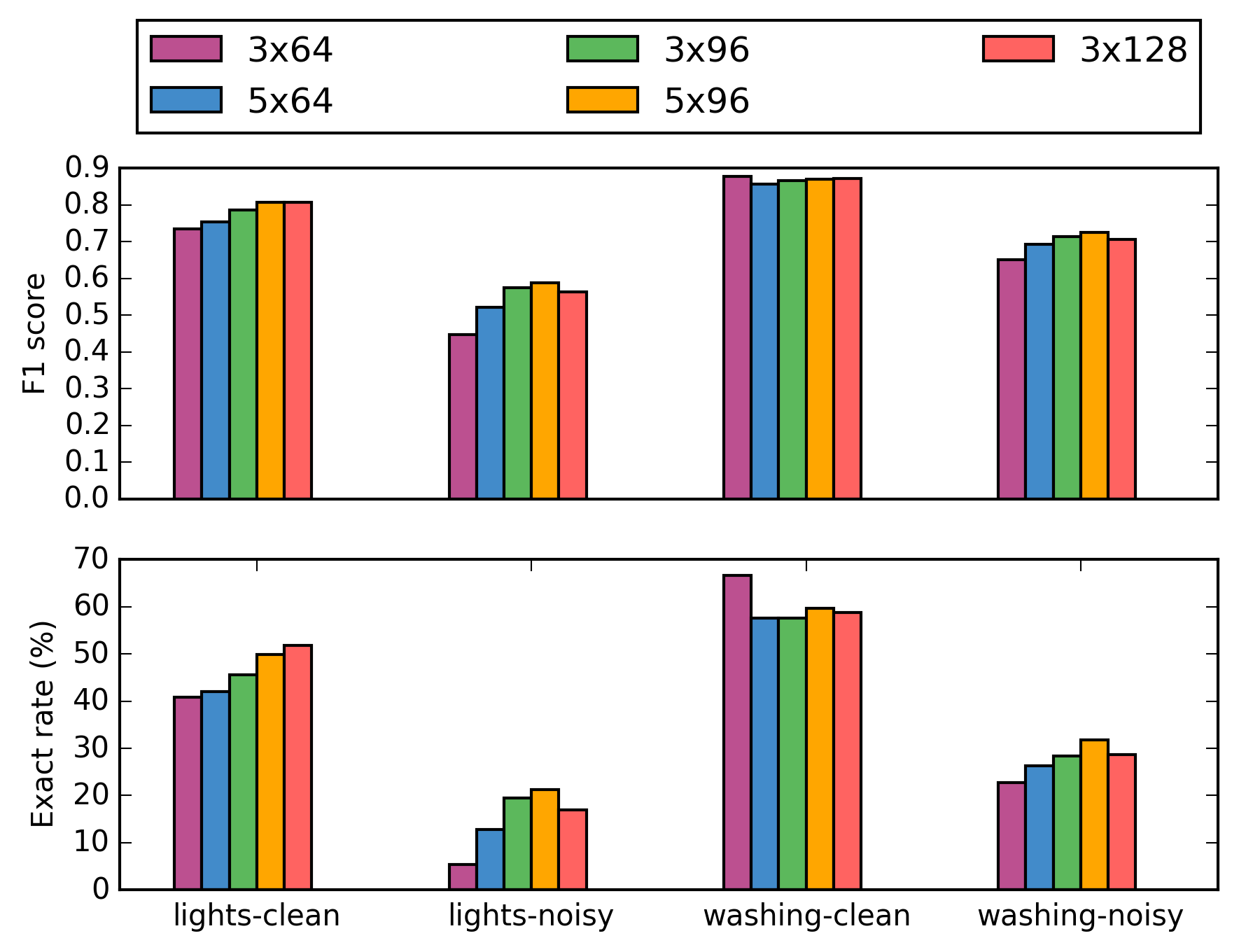}
    \caption{Comparison of F1-scores and exact rates for \textbf{different sizes of quantized networks}.}
    \label{fig:final_nnet_size}
\end{figure}

We show in Fig.~\ref{fig:final_nnet_size} the performance of quantized networks of different sizes, using the sequence post-processor and ``\textit{no-blank}'' confidence ($C_{nb}$). With some exceptions, bigger models are better. The best neural network across the different datasets seems to be the \texttt{5x96} architecture, with $395k$ parameters.

\begin{figure}[!t]
    \centering
    \includegraphics[width=\linewidth]{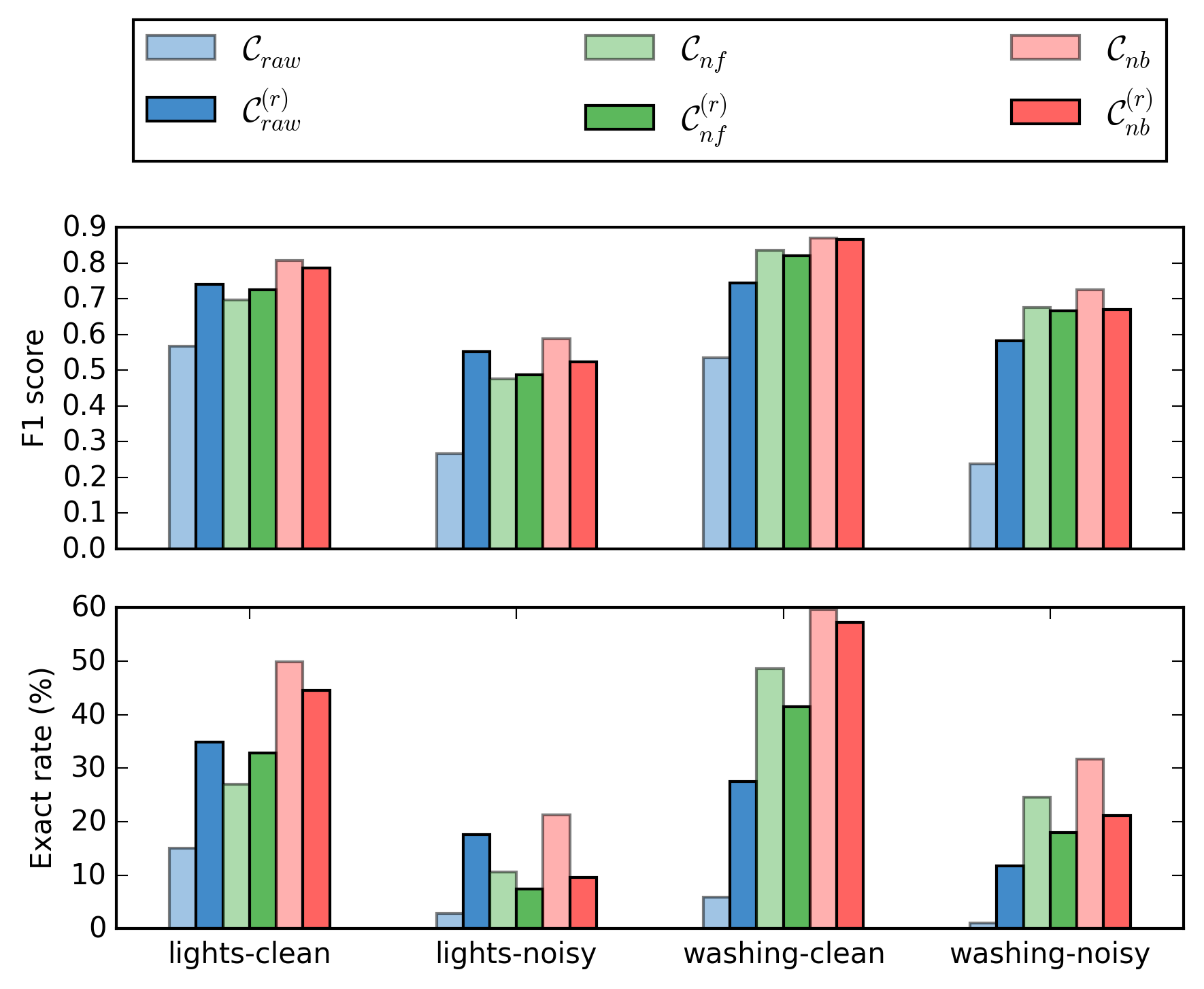}
    \caption{Comparison of F1-scores and exact rates for the quantized \texttt{5x96} architecture for \textbf{different confidence scores}.}
    \label{fig:final_conf}
\end{figure}

In Fig.~\ref{fig:final_conf}, we compare different choices of confidence scores for that network. We see that the raw confidence score ($C_{raw}$) gives the poorest performance. Normalizing by the segment length gives much better results ($C_{nf}$). The normalization by the estimated number of blank frames ($C_{nb}$) yields the best results for both metrics and all four datasets. Using likelihood (or confidence) ratio helps a lot for the raw confidence, but always gives worse results for ``\textit{no-blank}''. For $C_{nf}$, no clear conclusion can be drawn.

\begin{figure*}[!t]
    \centering
    \includegraphics[width=0.8\linewidth]{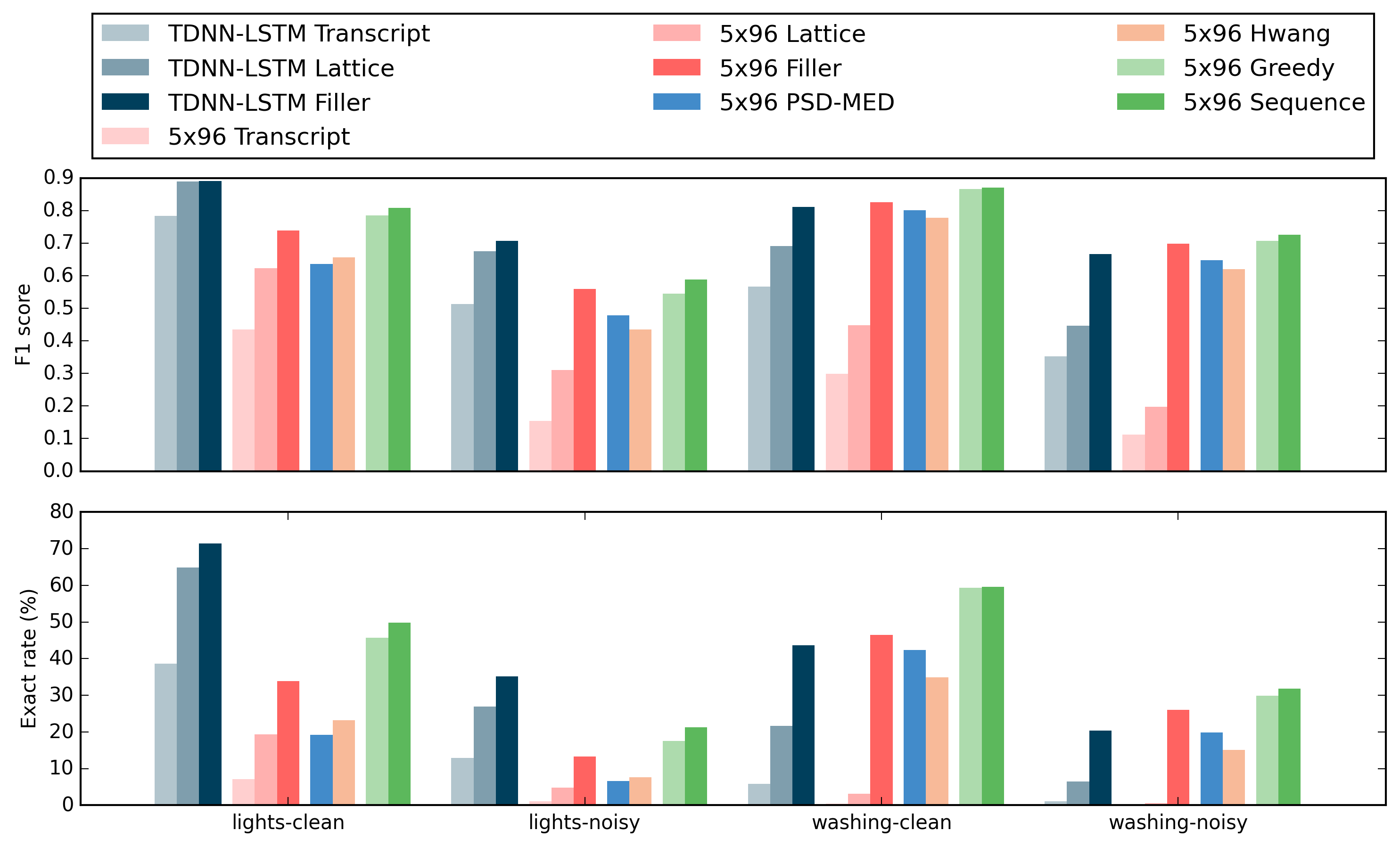}
    \caption{Comparison of F1-scores and exact rates for the \texttt{5x96} architecture and the large TDNN-LSTM for \textbf{different KWS methods}.}
    \label{fig:final_postproc}
\end{figure*}

\begin{table}[!t]
\begin{center}
 \begin{tabular}{lrcccc}
 & & \multicolumn{2}{c}{\textbf{lights}} & \multicolumn{2}{c}{\textbf{washing}} \\
Model & Method   & clean & noisy & clean & noisy \\ \hline
 TDNN-LSTM  & Transcript & 0.783 & 0.513 & 0.567 & 0.352 \\
            & Lattice & 0.889 & 0.674 & 0.692 & 0.446 \\
            & Filler & 0.891 & 0.707 & 0.811 & 0.666 \\\hline
 5x96  & Transcript & 0.435 & 0.154 & 0.299 & 0.113 \\
 (quantized)      & Lattice & 0.623 & 0.311 & 0.448 & 0.198 \\
       & Filler & 0.739 & 0.560 & 0.826 & 0.698 \\\cline{2-6}
       & PSD-MED & 0.636 & 0.478 & 0.800 & 0.647 \\
       & Hwang & 0.657 & 0.435 & 0.778 & 0.620 \\\hline
 \multicolumn{2}{r}{Proposed (greedy)}  & 0.784 & 0.545 & 0.866 & 0.708 \\
 \multicolumn{2}{r}{Proposed (sequence)}  & \textbf{0.808} & \textbf{0.588} & \textbf{0.871} & \textbf{0.725} \\\hline

\end{tabular}   
\end{center}
\caption{Comparison of F1-scores of the quantized \texttt{5x96} architecture for different post-processors.}
    \label{tab:final}
\end{table}

Finally, we compare the proposed method to the baselines. The exact rates and F1 scores of all methods are shown in Fig.~\ref{fig:final_postproc} and the F1 scores are also reported in Table~\ref{tab:final}. The proposed method is configured with the quantized 5x96 architecture and the \textit{no-blank} confidence without ratio ($C_{nb}$). We used that same network for different baseline methods, and also compare a few methods with the large TDNN-LSTM network.

We notice that the \textit{``Transcript''} LVCSR baseline performs poorly. This may be expected from the word error rates presented in Table~\ref{tab:asr_slu}. A big improvement over this baseline is achieved by using lattices instead of the Viterbi decoding. Using a filler model instead of an LVCSR approach provides further improvement, especially for the small network. The PSD-MED and Hwang methods are a bit worse than the filler model, which is consistent with previous publications~\cite{chenthesis}, and may also be explained by the lack of word boundary class in our model compared to the reference ones presented in the corresponding papers~\cite{zhuang2016unrestricted,chen2018sequence,hwang2015online}. 

The method we propose in this paper, in green in Fig.~\ref{fig:final_postproc}, performs better than the filler method using the same neural network. The sequence post-processor gives a small improvement over the greedy approach. It even reaches better results than the filler model using the TDNN-LSTM model on the ``washing machine'' dataset. Although the results are a bit behind on the ``smart light'' dataset, it is nonetheless better than the LVCSR ``transcript'' approach, with a much smaller network and fast decoding.


\section{Conclusion}

We have presented a small-footprint keyword spotting method that does not require training data labeled at the keyword level, applied to a spoken language understanding scenario in which multiple keywords should be detected in a single query. The acoustic neural network is trained with CTC on generic ASR data, and can be used to detect any arbitrary keyword. We proposed a quantization scheme for LSTM layers, allowing us to build a small neural network weighing less than 500KB, which can run in real-time on micro-controllers. 

We exploited the characteristics of the typical outputs of a CTC-trained network to optimize the keyword spotting algorithm. We proposed a confidence score calibration based on a normalization of the CTC score by an estimate of the number of \textit{``blank''} frames, and gained a factor two in the speed of the detection algorithm by dropping the blank frames. Moreover, we carried out a comprehensive exploration of the impact of different aspects of the proposed system. 

We compared the detection performance of our method to standard baselines, either based on a LVCSR system or on a filler model. We have shown that our approach outperforms the filler model on the studied tasks and datasets, as well as several other approaches proposed in the literature.

Future work will focus on developing the SLU system on top of the outputs of this models, potentially providing cues to improve the decoding of keyword sequences. We will also use that system as a baseline to evaluate our future work on open-vocabulary KWS models.



\bibliographystyle{IEEEtran}
\bibliography{paper}
\vfill

%





\end{document}